\documentclass{article}
\usepackage{arxiv}

\title{Approximate Inference for Stochastic Planning in Factored Spaces}
\author[ ]{Zhennan Wu}
\author[ ]{Roni Khardon}

\affil[ ]{%
    Department of Computer Science\\
    Luddy School of Informatics, Computing, and Engineering\\
    Indiana University Bloomington\\
    Bloomington, Indiana, USA
}
\affil[ ]{\textit {\{zwu1, rkhardon\}@iu.edu}}
 \begin{document}
\maketitle

\begin{abstract}
Stochastic planning can be reduced to probabilistic inference in large discrete graphical models, but hardness of inference requires approximation schemes to be used. In this paper we argue that such applications can be disentangled along two dimensions. The first is the direction of information flow in the idealized exact optimization objective, i.e., forward vs.\  backward inference. 
The second is the type of approximation used to compute this objective, e.g., Belief Propagation (BP) vs.\ mean field variational inference (MFVI). 
This new categorization allows us to unify a large amount of isolated efforts in prior work 
explaining their connections and differences as well as potential improvements. 
An extensive experimental evaluation over large stochastic planning problems shows the advantage of forward BP over several algorithms based on MFVI. An analysis of practical limitations of MFVI motivates a novel algorithm, collapsed state variational inference (CSVI), which provides a tighter approximation and achieves comparable planning performance with forward BP. 
\end{abstract}

\section{Introduction}\label{sec:intro}
The connection between planning and probabilistic inference is well known and multiple reductions exist 
showing how inference algorithms can be used to solve stochastic planning problems. 
Such reductions are equivalent when one can perform exact inference but this is not typically the case for challenging planning problems that have many state variables, a.k.a.\ factored spaces, where approximate inference schemes are introduced. The planning and reinforcement learning literatures include multiple such efforts where different algorithmic frameworks are combined with different approximation schemes. 
For example, constructions exist through 
weighted model counting \citet{domshlak2006fast}, 
several forms of variational inference
(e.g., \citep{toussaint2006probabilistic,levine2018reinforcement}),
and several forms belief propagation 
(e.g., \citep{liu2012belief,cui2019stochastic}).
However, it is not clear how different algorithmic approaches are related to one another  and how the choice of approach interacts with the choice of approximation scheme.

The paper makes three contributions. First we provide a unified scheme that connects previous approaches along 
two dimensions,
using either forward 
or backward reasoning, and choosing 
what approximation to use,
where we address Belief Propagation (BP) and mean field variational inference (MFVI).
This allows us to put prior work 
in a unified framework that explains choices made by corresponding algorithms. 
In particular, our analysis shows that Forward MFVI which is used in some papers can be understood to run multiple iterations of Backward MFVI and thus provides tighter approximations.
Second, through extensive experiments over large planning problems, we show that forward reasoning with Belief Propagation provides the best performance among these algorithms, that MFVI provides poor performance in some domains, and that modifying MFVI using exponentiated rewards helps in some cases but not sufficiently. 
We also analyze the failures of MFVI experimentally pointing to sensitivity in updates. 
Third, based on the analysis, we propose a novel algorithm, Collapsed State Variational Inference (CSVI), that uses mean field with collapsed variational inference where state variables are integrated out. CSVI is motivated theoretically due to its tighter variational approximation and we show empirically that it matches the performance of Forward Belief Propagation. 
This shows that while naive application of mean field for planning fails, 
other variational approximations like CSVI can yield strong planning performance.
Due to space constraints some technical details and experiment results are omitted from the main paper and are provided in \citep{WuKh2022arxiv}, which we refer to below as the \reftosupp.

\section{Problem Formulation}

We consider finite horizon MDPs as specifications of planning problems. Such specifications are often compiled from a high level description language 
but this is orthogonal to the discussion in the paper. 
Specifically, consider Markov Decision Processes $\langle \mathcal{S}, p(s_0), \mathcal{A}, \mathcal{P},  \mathcal{R}, T, \gamma \rangle$, where $\mathcal{S}$ denotes the state space, 
$p(s_0)$ is a distribution over start states,
$\mathcal{A}$ denotes the action space, $\mathcal{P}$ denotes the transition probability $p(s_{t+1}|s_{t}, a_{t})$, $\mathcal{R}$ denotes the reward function $R(s_{t}, a_{t})$, $T$ denotes the horizon and $\gamma$ is the discount factor. 
In this paper we set $\gamma=1$, 
but this does not significantly affect any of the formulations. 
A solution is given by a policy, $\pi$, that specifies $p_{\theta_t}(a_t|s_t)$ with policy parameters
$\theta=\{\theta_t\}$ allowing for non-stationary policies. 
The task in planning is to find a policy that maximizes the expected cumulative reward
$\mathbb{E}[\sum_{t=0}^{T-1}R(s_t, a_t)]$
where the expectation is taken w.r.t.\ trajectories generated from the MDP with policy $\pi$, that is, $s_0\sim p(s_0)$, $a_t\sim p_{\theta_t}(a_t|s_t)$, and $s_t\sim p(s_t|s_{t-1},a_{t-1})$.

In this paper we follow recent practice in stochastic planning and use the {\em online planning} framework, where in a state $s$, the algorithm computes for a limited time to pick an action $a$, uses $a$ to control the MDP to get to the next state, and repeats this process. Online planning is often used with receding horizon control, where the planner uses a $T$ step lookahead in its search and then extracts the first action $a$ to be applied in $s$.

Solving a finite horizon MDP is equivalent to solving an inference problem in the corresponding Dynamic Bayesian Network (DBN), or more precisely in the dynamic decision network.
We assume a factored form of states consisting of binary state variables, i.e. $s_t = (s_t^1, \cdots, s_t^M)$. We also assume a factored action representation $a_t = (a_t^1, \cdots, a_t^N)$.

The MDP formulation above requires real-value reward nodes in the DBN.
To facilitate inference one can replace these nodes with constructions that use only binary variables, and various such constructions appear in the literature. In the following we develop one such construction and use that in  our experiments. 
We introduce binary reward random variables $r_t$ to capture the reward after taking action in the previous time step $t-1$, the distribution of which is defined as 
\begin{align}
    p(r_t = 1|s_{t-1} = s, a_{t-1} = a) = \frac{R(s_{t-1} = s, a_{t-1} = a)}{\max_{s, a}R(s, a)}. 
\end{align}
We can then define $\tilde{R}$ where $p(\tilde{R}=1)=\frac{\sum_1^T r_t}{T}$ to capture the cumulative reward. We call the resulting DBN the intermediate representation. 
However, $\tilde{R}$ has $T$ parents which hinders efficient inference. 
To avoid the use of $\tilde{R}$, 
we introduce cumulative reward binary random variables $c_t$. To keep the consistency of the graphical structure, we create an auxiliary node $c_0 \equiv 1$, and for $t > 0$ the distribution of $c_t$ is defined recursively depending on the previous cumulative reward $c_{t-1}$ and current reward $r_t$:
\begin{equation}\label{form}
    p(c_t = 1|c_{t-1}, r_t) = \frac{(t-1)c_{t-1} + r_t}{t}.
\end{equation}
In many planning problems the reward is given as an additive function over a set of small factors. 
For such problems we introduce another chain of binary reward variables within a time step
using a similar construction.
This yields a DBN that only includes binary variables with a small number of parents.
As the following proposition shows the three constructions, using cumulative reward, using $\tilde{R}$ and using $c_T$ are equivalent. Further details and proofs are given in the \reftosupp.
\begin{proposition}
The construction satisfies
    $\mathbb{E}[\sum_{t=0}^{T-1}R(s_t, a_t)] \propto  \mathbb{E}(\tilde{R}) \propto \mathbb{E}(c_T)
    $
where expectations are w.r.t.\ trajectories as above.
\end{proposition}


\section{Planning Through Inference}
In the following we restrict our discussion to open loop policies, that is, $p_{\theta_t}(a_t|s_t)=p_{\theta_t}(a_t)$ where the policy is time dependent but does not depend on the state (other than $s_0$ if it is fixed). 
Thus for an action sequence $A=\{a_0,\ldots,a_{T-1}\}$, we have $p_\theta(A)=\prod p_{\theta_t}(a_t)$.
This covers most of previous work on planning as inference in the literature. The extension to standard policies is straightforward but requires more complex algorithms for optimization.

\subsection{Forward Backward Framework} 
We now present a simple framework that captures many algorithms in the literature.
For the discussion below note that some algorithms optimize policy parameters ${\theta}$ and then choose the actions, whereas others optimize the action sequence $A$ directly.

\noindent
\textbf{The Backward Framework:} Observe that
if $\theta$ is the uniform distribution, $u$, then 
    \begin{equation}
        \argmax_A p(c_T = 1|A) = \argmax_A\frac{p_{u}(A|c_T = 1)p_u(c_T = 1)}{p_u(A)} = \argmax_A p_u(A|c_T = 1)    
    \end{equation}
    where the second equality is true because $p_u(A)$ is a fixed constant for all $A$ and $p_u(c_T = 1)$ does not depend on $A$.
This suggests that we can optimize $p(c_T = 1|A)$ by optimizing $p_u(A|c_T = 1)$.
Since calculating $p_u(A|c_T = 1)$ is hard, 
the backward framework optimizes an approximation of $p_u(A|c_T = 1)$.
The choice of different approximations $q_{\phi}(A)$ will give us different concrete algorithms.
This is captured in Algorithm~\ref{alg:backward}
\begin{algorithm}
    \caption{Backward Inference}
    \begin{algorithmic}
        \State 1. Calculate $q_{\phi}(A) \approx p_u(A|c_T = 1)$
        \State 2. Pick $A = \argmax q_{\phi}(A)$
    \end{algorithmic}
\label{alg:backward}
\end{algorithm}

\noindent
\textbf{The Forward Framework:} in contrast, the forward approach aims to directly optimize $p_\theta(c_T=1)$ w.r.t the policy parameters (or alternatively, $p(c_T=1|A)$ but we focus on the more general case).
Approximating $p_\theta(c_T=1)$ with a score function $sc(\theta)$ defined on policy parameters yields the forward framework. 
In the ideal case, maximizing $sc(\theta)$ will give us a delta function, directly  selecting a concrete $A$ sequence. If not, we can use $\argmax$ or sample from the corresponding distribution. 
This is captured in Algorithm~\ref{alg:forward}

\begin{algorithm}
    \caption{Forward Inference}
    \begin{algorithmic}
        \State 1. 
        Define a score function 
               $sc(\theta) \triangleq sc(c_T = 1|\theta) \approx p_{\theta}(c_T = 1)$
        \State 2. Optimize $\theta$ to maximize the score function.
        \State 3. Pick $A$ using $p_{\theta}(A)$
    \end{algorithmic}
\label{alg:forward}
\end{algorithm}

\subsection{Forward and Backward Loopy Belief Propagation}
The forward and backward algorithms can be combined with any approximation scheme. 
We start by considering loopy BP (LBP) algorithms \citep{pearl1988probabilistic,Kschischang2001}.
For this construction we translate the DBN into a factor graph using standard constructions.
For backward LBP, we
instantiate $c_T=1$ as evidence, 
fix the factors corresponding to $\theta$ to be the uniform distribution,
and run LBP to calculate the marginal probabilities on action variables. 
That is, $q_\phi(A)$ is given by the output of LBP. 
Note that this is algorithmically simple  because we do not need a separate optimization step aside from Belief Propagation. However, LBP may need many iterations to converge or may not converge at all.

For the forward algorithm, we define $sc(\theta)$ to be the approximate marginal of $p_{\theta}(c_T)$ computed by LBP. However, LBP does not optimize $\theta$. 
As discussed below, multiple techniques for optimizing $\theta$ for LBP exist in the literature. 
In the experiments
we use the SOGBOFA system \citep{cui2019stochastic} that combines LBP with gradient based search. 



\subsection{Forward and Backward Mean Field Variational Inference}
The idea in variational inference is to minimize the KL divergence between the approximate posterior and the true posterior over latent variables, i.e., in our case 
\begin{align*}
d_{KL}(q_{\phi}(S, A, R, C_{\backslash T}) || p_{\theta}(S, A, R, C_{\backslash T}| c_T = 1))
\end{align*}
where
the latent variables are $S$, $A$, $R$, $C$, that is, the sequences of state, action, reward, and cumulative reward variables, where $C_{\backslash T}$ excludes $c_T$. 
This is equivalent to maximizing the {evidence lower bound (ELBO)}.
In our case the ELBO is given in the next equation, where
in the mean field approximation $q_{\phi}$ is a product of independent factors
\begin{equation}
    \log p_{\theta}(c_T = 1) \geq \mathbb{E}_{q_{\phi}}[\log \frac{p_{\theta}(S, A, R, C_{\backslash T}, c_T = 1)}{q_{\phi}(S, A, R, C_{\backslash T})}] =: ELBO_{\theta, \phi}.
\end{equation}

For backward MFVI, note that $p_u(A|c_T = 1)$ is the marginal distribution of the true posterior $p_{u}(S, A, R, C_{\backslash T}|c_T = 1)$. Therefore we first 
maximize $ELBO_{\phi, \theta = u}$ to obtain 
$q_{\phi}(S, A, R, C_{\backslash T})$ and then set
$q_{\phi}(A)$ to be the corresponding marginal.
Detailed update equations for MFVI are given in the \reftosupp .

For forward MFVI, we can pick $sc(\theta) = ELBO_{\phi, \theta} \approx \log p_{\theta}(c_T = 1)$
where we need to optimize both $\phi$ and $\theta$. For this, the standard approach is the Variational Expectation Maximization 
algorithm which optimizes $\phi$ in the $E$ step and $\theta$ in the $M$ step.
To elaborate the algorithm, note that the ELBO can be reformulated as follows:
\begin{align}
        ELBO_{\theta, \phi} = \mathbb{E}_{q_{\phi}}[\log \frac{p(S, R, C_{\backslash T}, c_T = 1|A)}{q_{\phi}(S, R, C_{\backslash T}|A)}] - d_{KL}(q_{\phi}(A)||p_{\theta}(A)) 
\label{eq:elbo2}
\end{align}
where
the first term does not depend on $\theta$.
Therefore:
\begin{itemize} 
\item 
In the E step, we maximize $ELBO_{\theta, \phi}$ w.r.t.\ $\phi$.
{\em  Note that this is exactly as in the Backward Algorithm but under a general $\theta$.}
 \item 
In the M-step, we keep $q_{\phi}$ fixed and optimize the $ELBO_{\theta, \phi}$ w.r.t. $\theta$.  
From Eq~(\ref{eq:elbo2})
we see that this is equivalent to minimizing $d_{KL}(q_{\phi}(A)||p_{\theta}(A))$. If $q_{\phi}(A)$ and $p_{\theta}(A)$ are from the same class of distributions, this step assigns $\theta \leftarrow \phi$. 
\end{itemize}
From the procedure, we have the following observation.
\begin{remark}
For the mean field approximation, the forward algorithm is an iterative process that alternates the backward algorithm with policy updates. 
\end{remark}
This connection was not observed in prior work where the forward and backward algorithms are not clearly distinguished.
Finally, as pointed by \citet{toussaint2006probabilistic} the E step is analogous to policy evaluation (except that we calculate marginals for many variables besides the reward) and the M step is analogous to policy improvement, so forward MFVI can be seen as an approximate version of Policy Iteration.

\section{Related Work}

The idea of using inference for stochastic planning has a long history and has attracted many different approaches. For example, \citet{Cooper1988} showed how inference can be used for decision making in influence diagrams,
\citet{domshlak2006fast} use an approach based on weighted model counting, 
\citet{nitti2015planning} use a probabilistic programming formulation, and  
\citet{Lee_Marinescu_Dechter_2021} use anytime marginal MAP solvers for planning problems. 

Several groups have developed approaches that follow the forward variational framework, going back to \citet{dayan1997using}.
This idea is often developed by defining a
reward weighted path distribution which is similar to conditioning on $c_T=1$ in our framework, and developing algorithms from this formulation 
\citep{furmston2010variational,furmston2012efficient,toussaint2006probabilistic,kumar2015probabilistic}.
We note, however, that these works did not explicitly address factoring over state and action variables.

On the other hand, some papers in robotics and reinforcement learning (RL)
 \citep{toussaint2009robot, kappen2012optimal,levine2018reinforcement}
follow the backward variational framework. In contrast with the discussion above they use a formulation where the reward  over trajectories is exponentiated. As shown by \citet{levine2018reinforcement}
this modifies the original optimization objective by adding a term with the expected entropy of the policy, 
and hence solves a slightly different problem, but the entropy term
may be beneficial for exploration in RL. 
In addition, the work of 
\citet{neumann2011variational} 
uses the forward variational algorithm, but with an exponentiated reward, and additional sampling-based approximations.
We can see that 
the forward and backward variational approaches have been widely used but have not been differentiated before.
Our analysis above clarifies the relationship between these approaches.

For the case of BP approximation, 
\citet{murphy2001factored} proposed the Factored Frontier Algorithm 
which is a forward BP method for marginal inference,
and 
\citet{BoyenKo1998} developed approximation bounds for forward inference. 
The work of 
\citet{liu2012belief,kiselev2014pomdp}
follows the forward BP framework, but
develops a generalized belief propagation algorithm 
that solves
both optimization and expectation steps using message passing.
The work of \citet{cui2019stochastic}
also follows the forward BP framework but
decouples the expectation which is done through BP from the optimization that uses an approximation based on gradient search. 

Several works have made additional assumptions on the structure of the DBN in their discussion of graph-based MDPs.
\citet{cheng2013variational} 
extend the algorithm of \citet{liu2012belief} to this case.
\citet{peyrard2006mean} and  \citet{sabbadin2012framework} use the Mean Field approximation method but only use it to approximate the distribution over state variables. They then use the approximate distribution to approximate steps of the Policy Iteration algorithm. Hence their algorithm is different from MFVI in that reward variables are not included in the variational approximation.   
Finally, our work can be seen to extend the comparison of 
Mean Field and Loopy BP
for general inference tasks
\citep{weiss2001comparing}. 
As in this early work, our experiments show that optimization of variational objectives
can lead to local optima and that BP can provide some advantage. 


\section{Experiments and Analysis of MFVI \& Belief Propagation Algorithms}\label{sec:floats}
\begin{figure*}[t]
	\centering
	\begin{subfigure}[t]{0.32\linewidth}
		\centering
		\includegraphics[width=\linewidth, trim=7 7 7 7, clip]{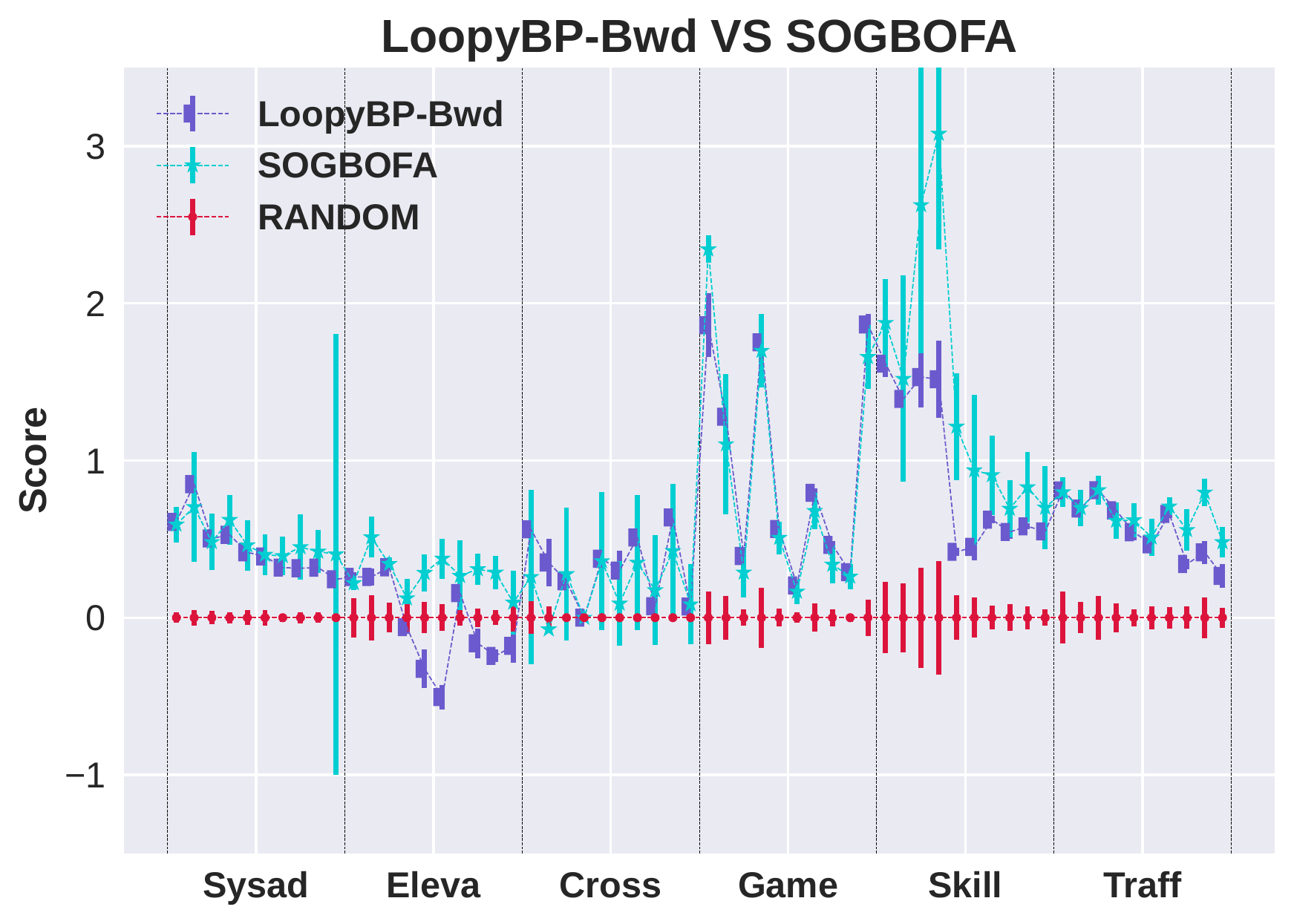}
	\end{subfigure}
    \begin{subfigure}[t]{0.32\linewidth}
    	\centering
    	\includegraphics[width=\linewidth, trim=7 7 7 7, clip]{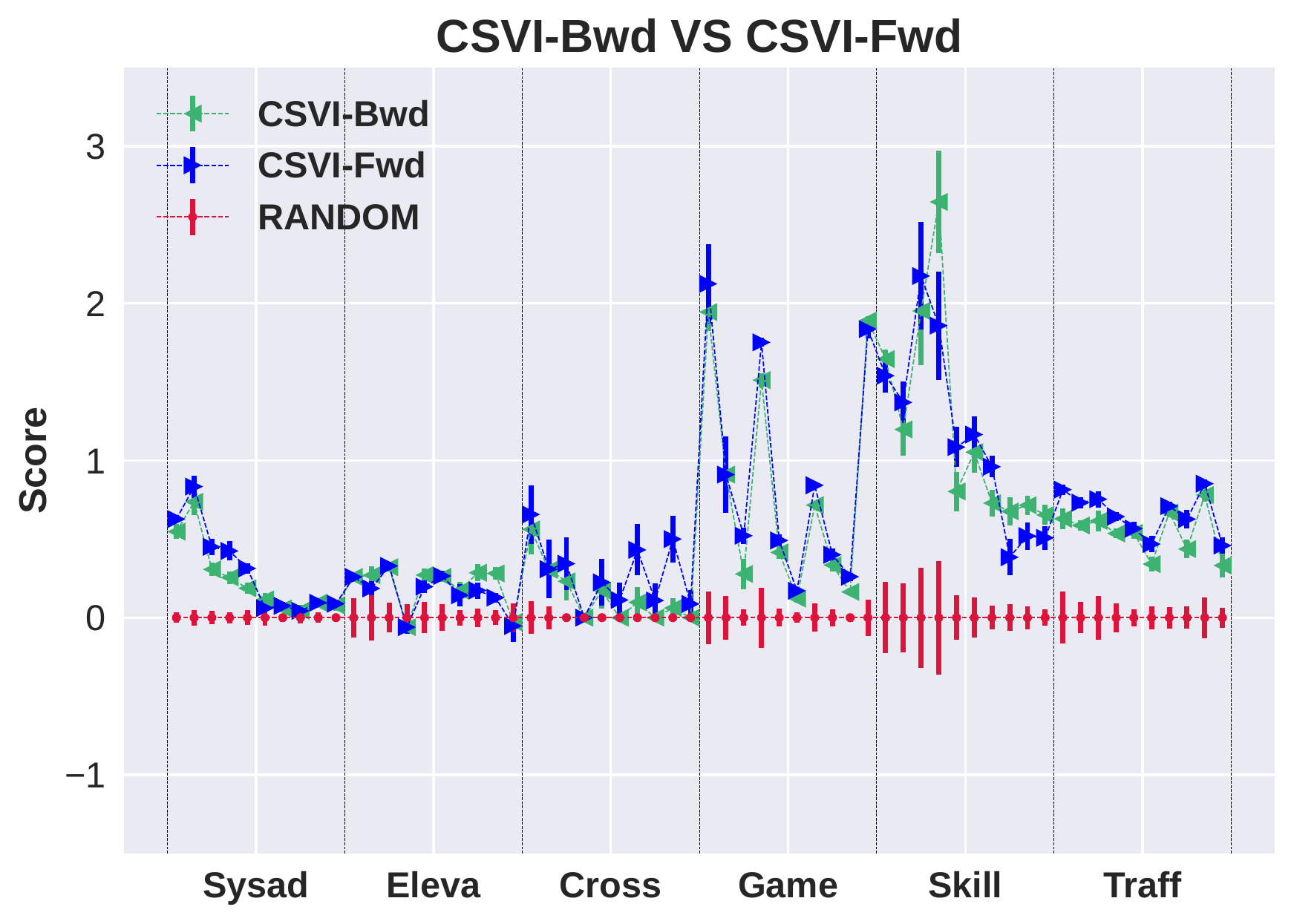}
    \end{subfigure}	
	\begin{subfigure}[t]{0.32\linewidth}
		\centering
		\includegraphics[width=\linewidth, trim=7 7 7 7, clip]{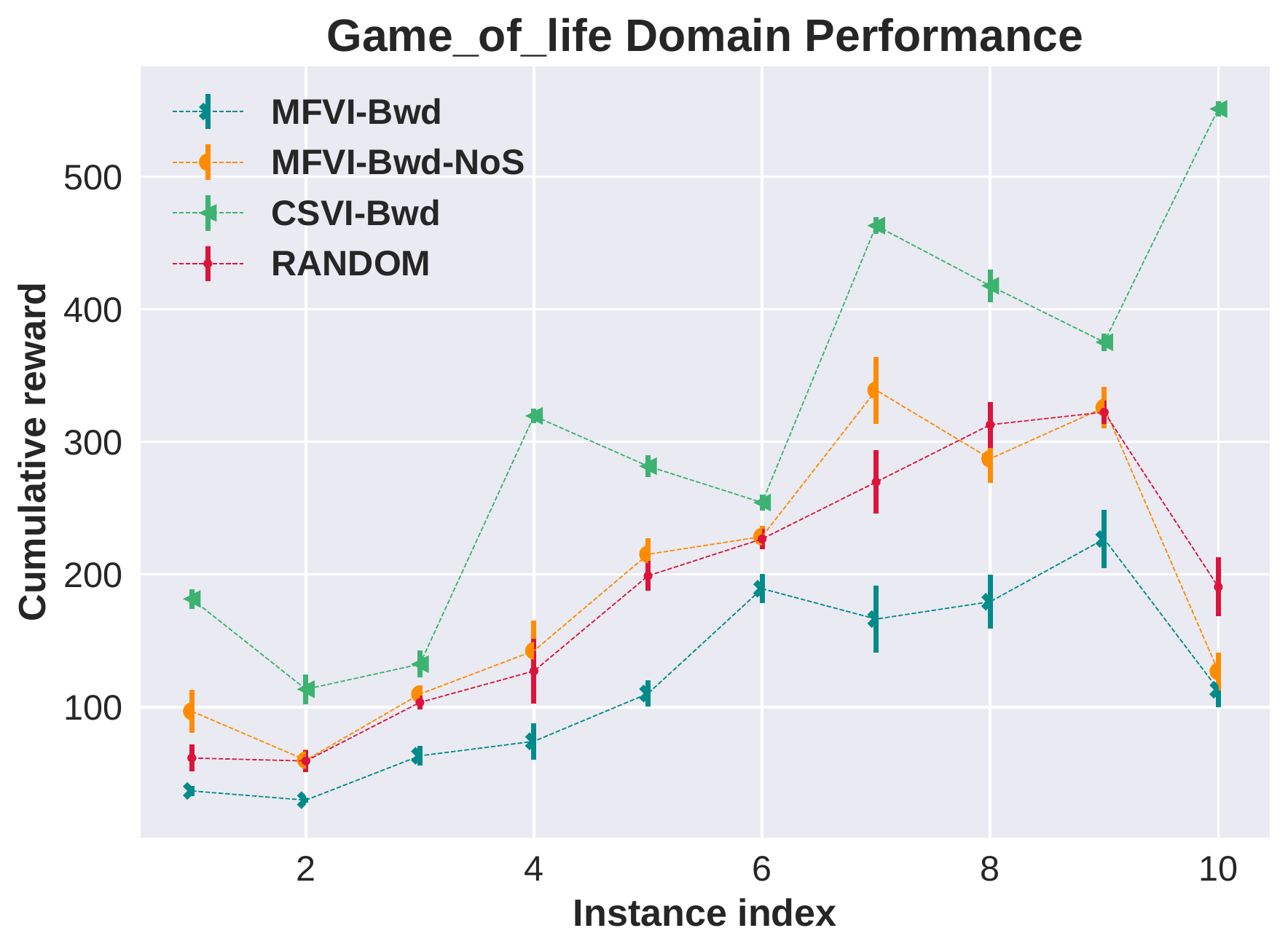}
	\end{subfigure}
\\
	\begin{subfigure}[t]{0.32\linewidth}
		\centering
		\includegraphics[width=\linewidth, trim=7 7 7 7, clip]{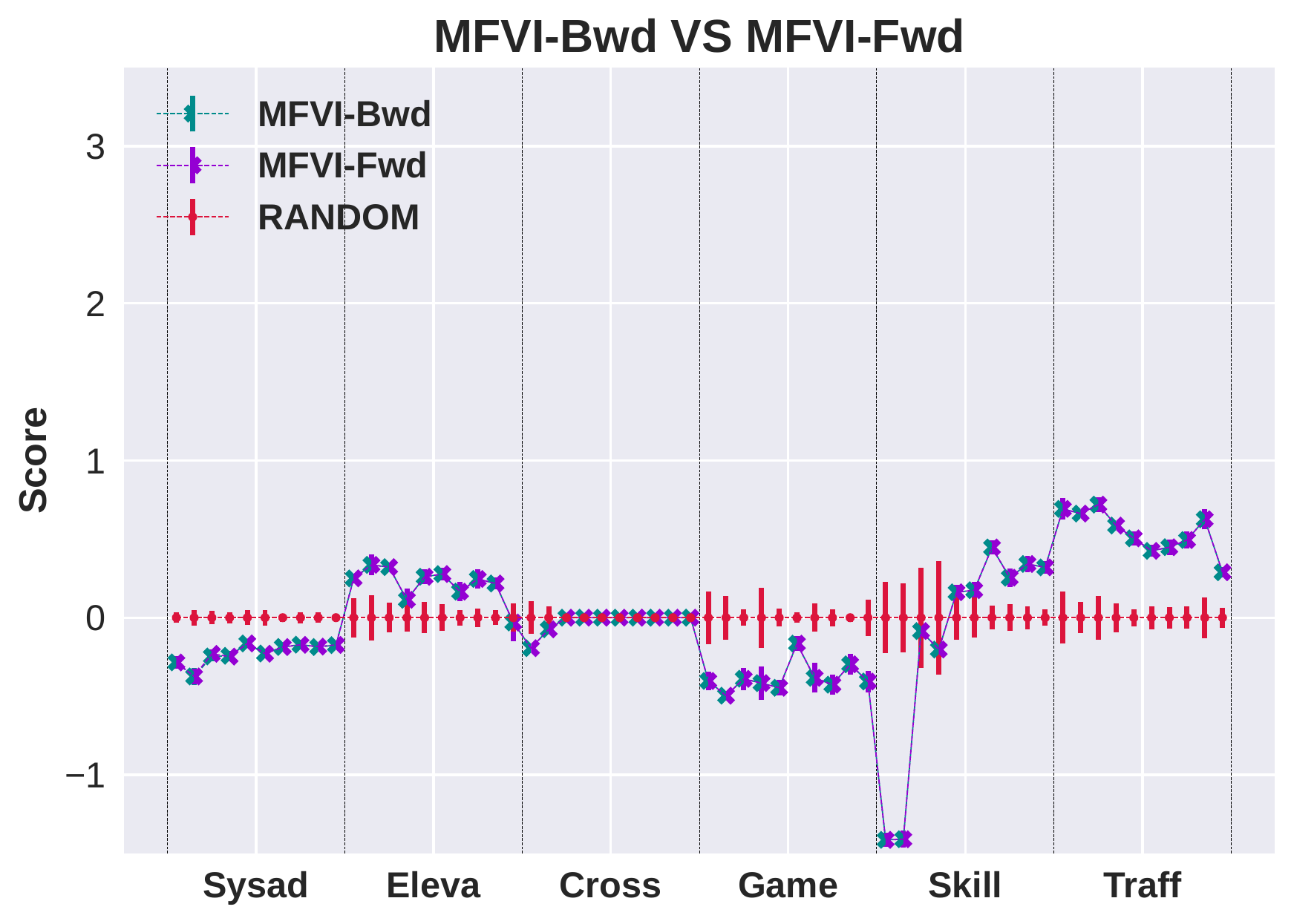}
	\end{subfigure} 
	\begin{subfigure}[t]{0.32\linewidth}
		\centering
		\includegraphics[width=\linewidth, trim=7 7 7 7, clip]{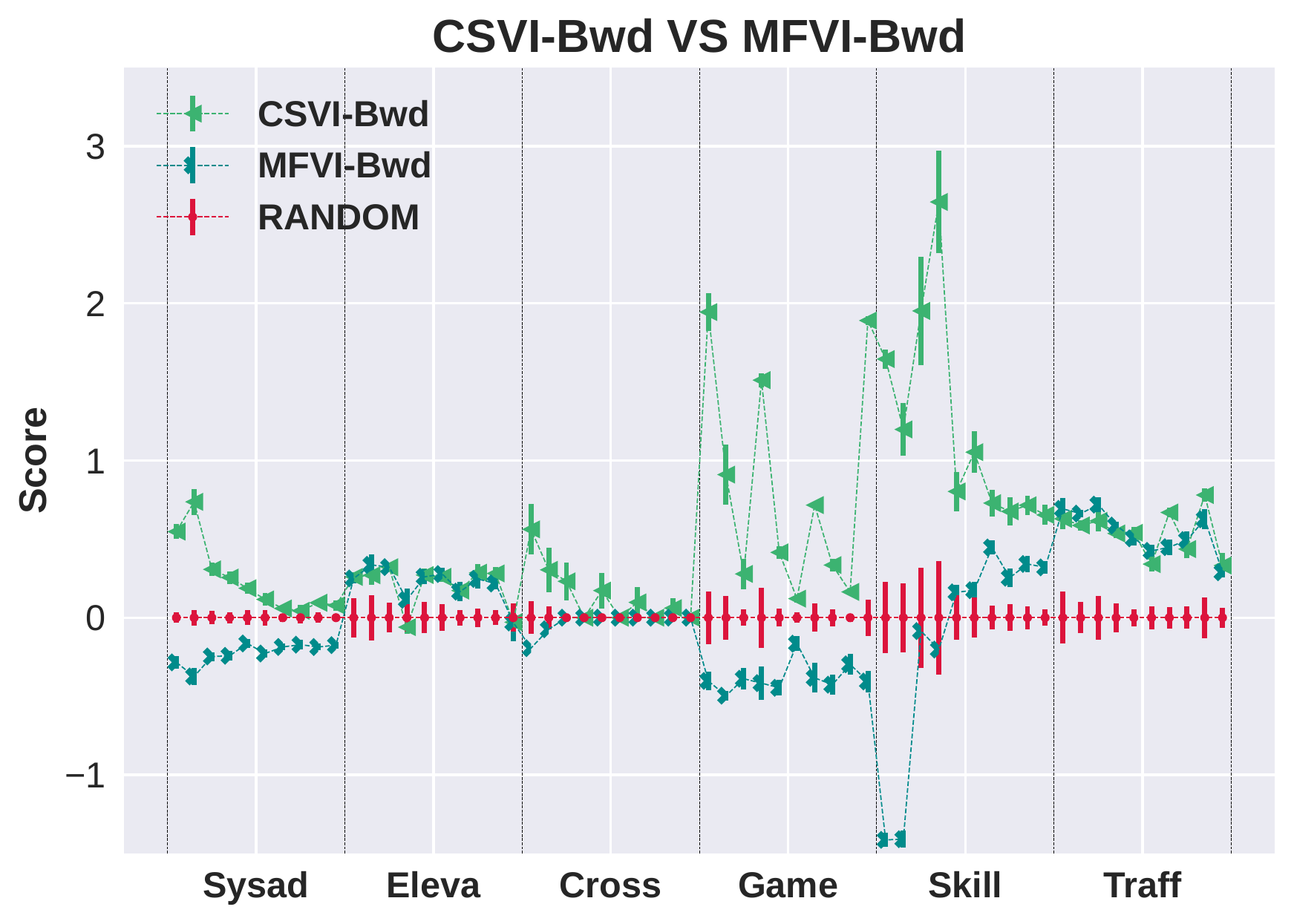}
	\end{subfigure}
	\begin{subfigure}[t]{0.32\linewidth}
		\centering
		\includegraphics[width=\linewidth, trim=7 7 7 7, clip]{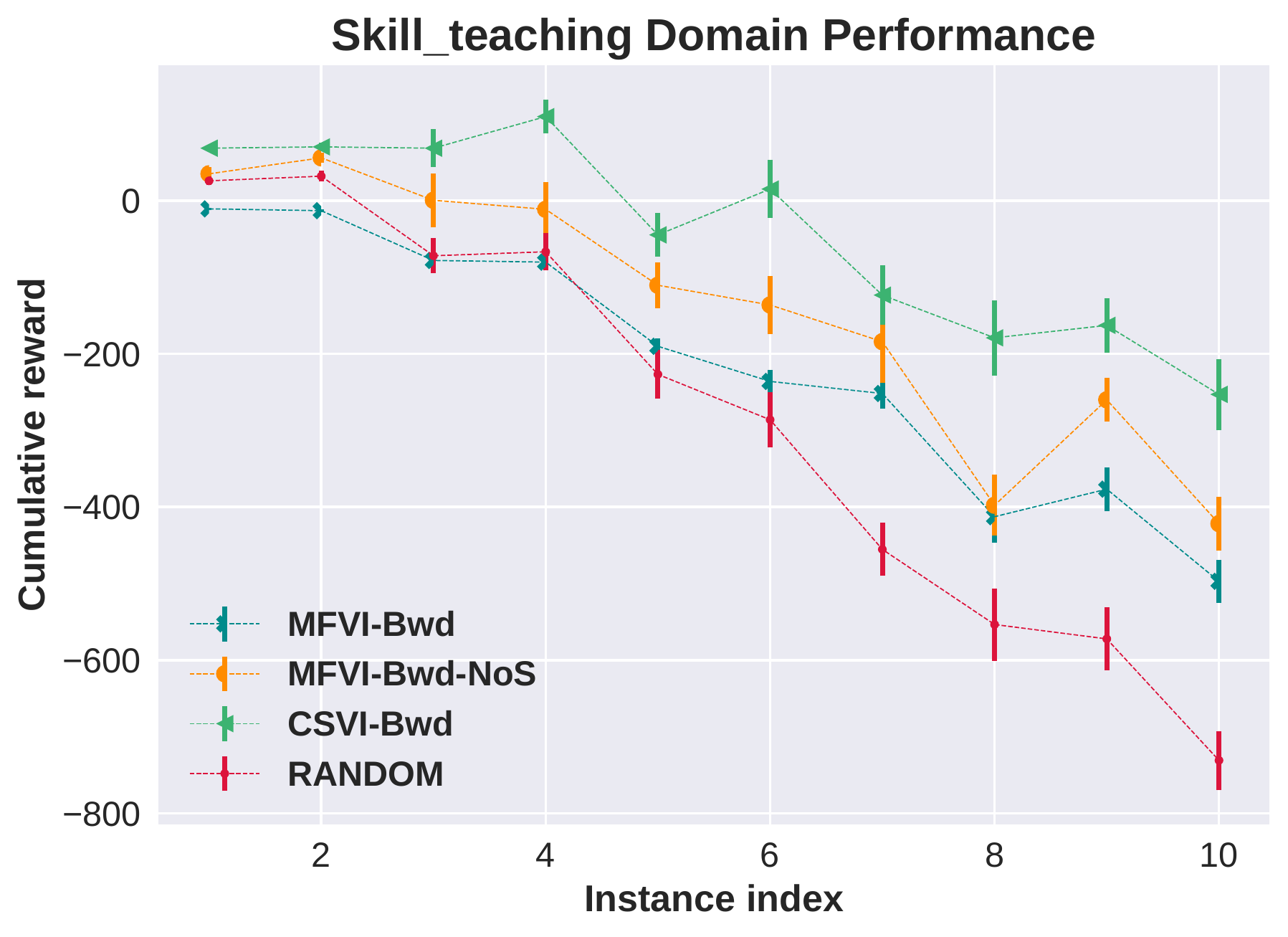}
	\end{subfigure}
\\
	\begin{subfigure}[t]{0.32\linewidth}
		\centering
		\includegraphics[width=\linewidth, trim=7 7 7 7, clip]{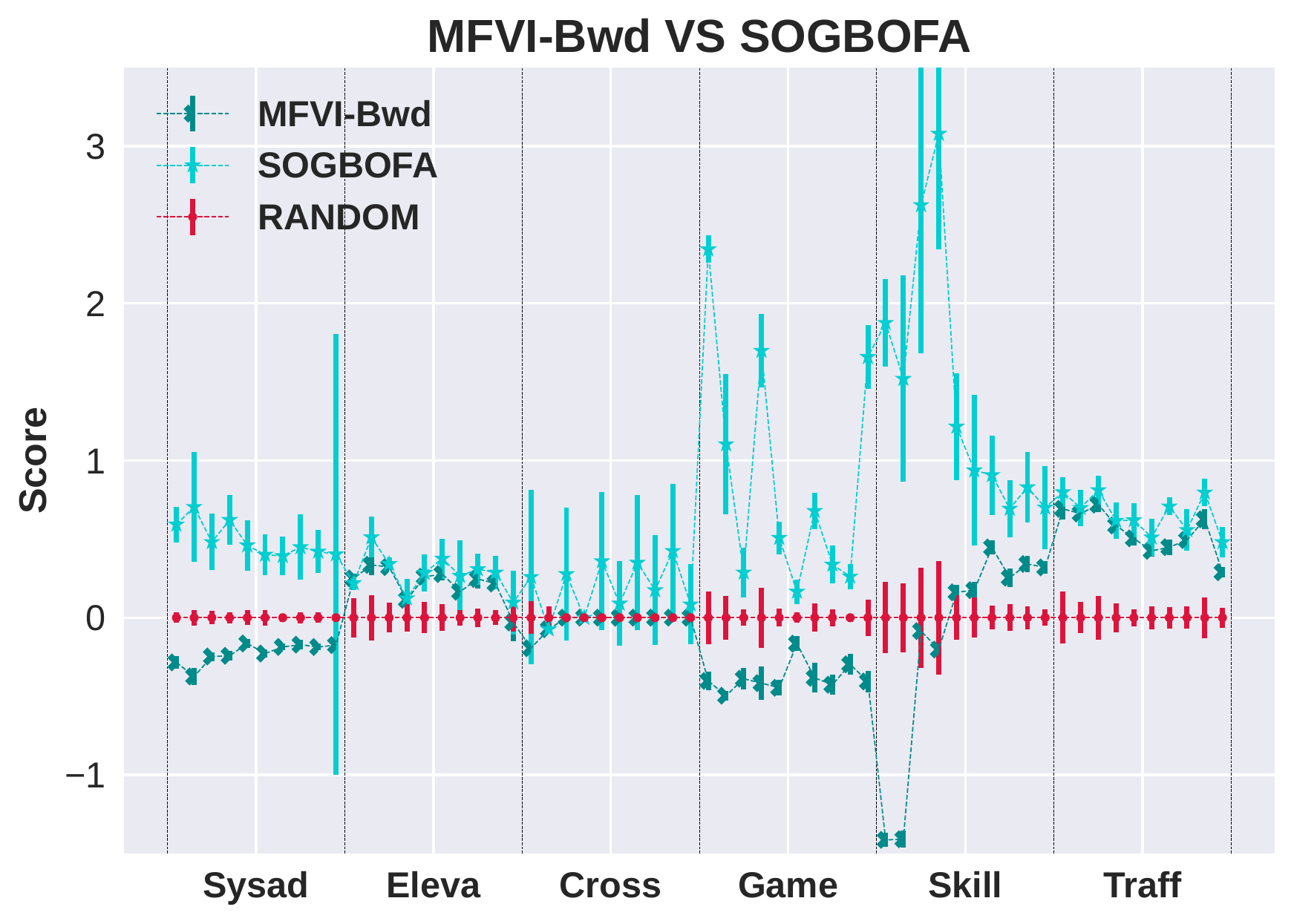}
	\end{subfigure}
	\begin{subfigure}[t]{0.32\linewidth}
		\centering
		\includegraphics[width=\linewidth, trim=7 7 7 7, clip]{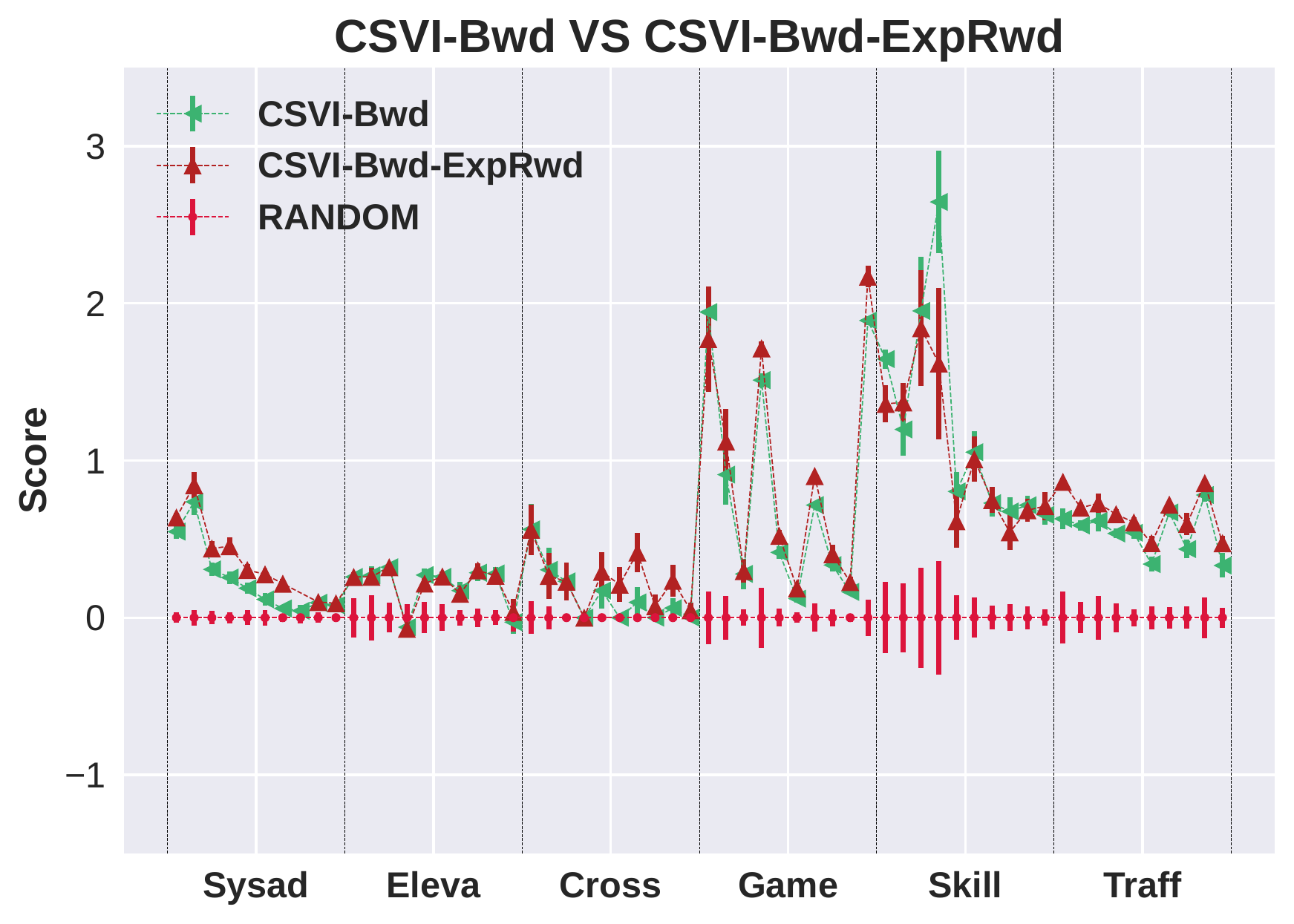}
	\end{subfigure}	
	\begin{subfigure}[t]{0.32\linewidth}
		\centering
		\includegraphics[width=\linewidth, trim=7 7 7 7, clip]{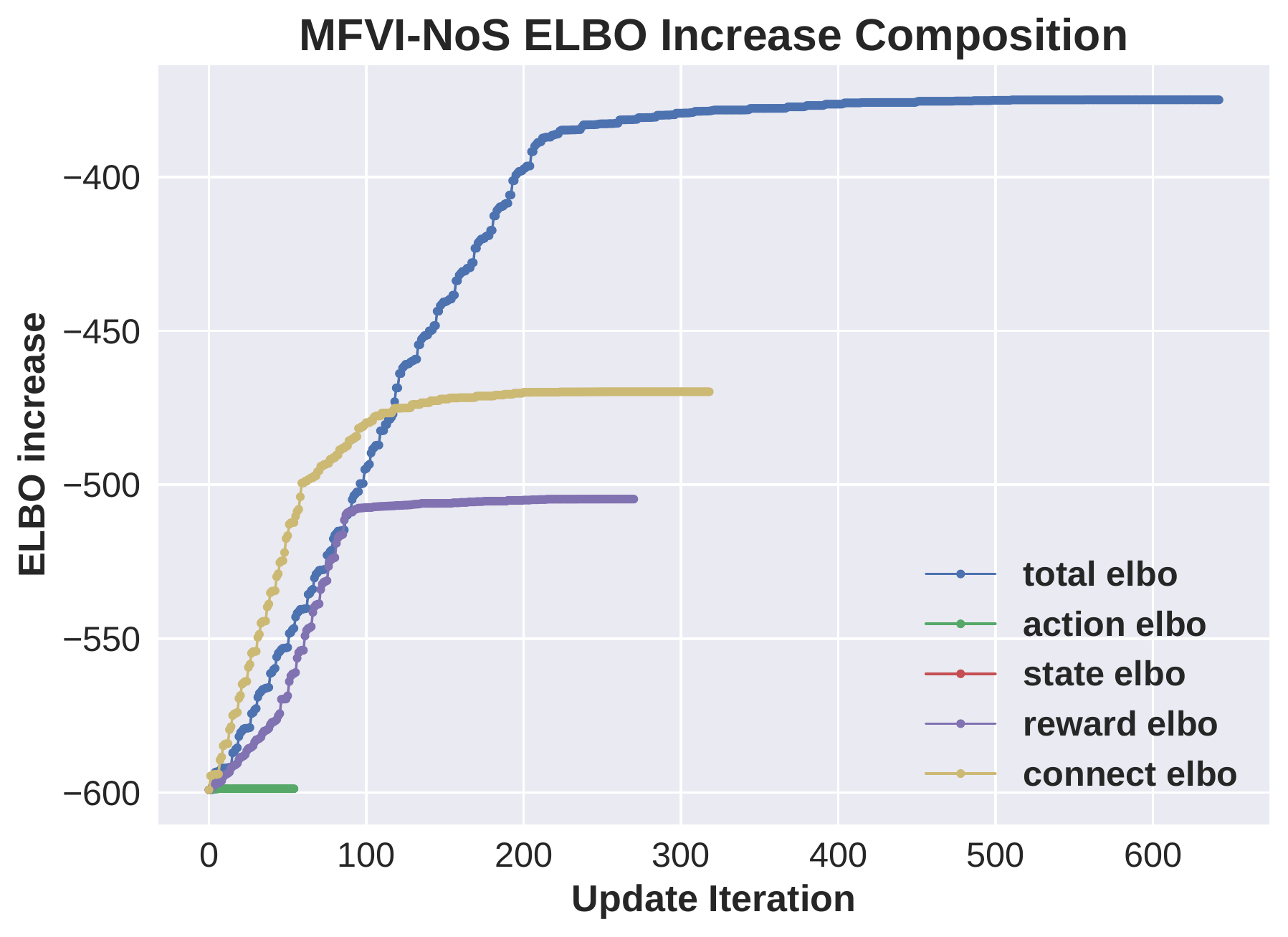}
	\end{subfigure}
\\
	\begin{subfigure}[t]{0.32\linewidth}
		\centering
		\includegraphics[width=\linewidth, trim=7 7 7 7, clip]{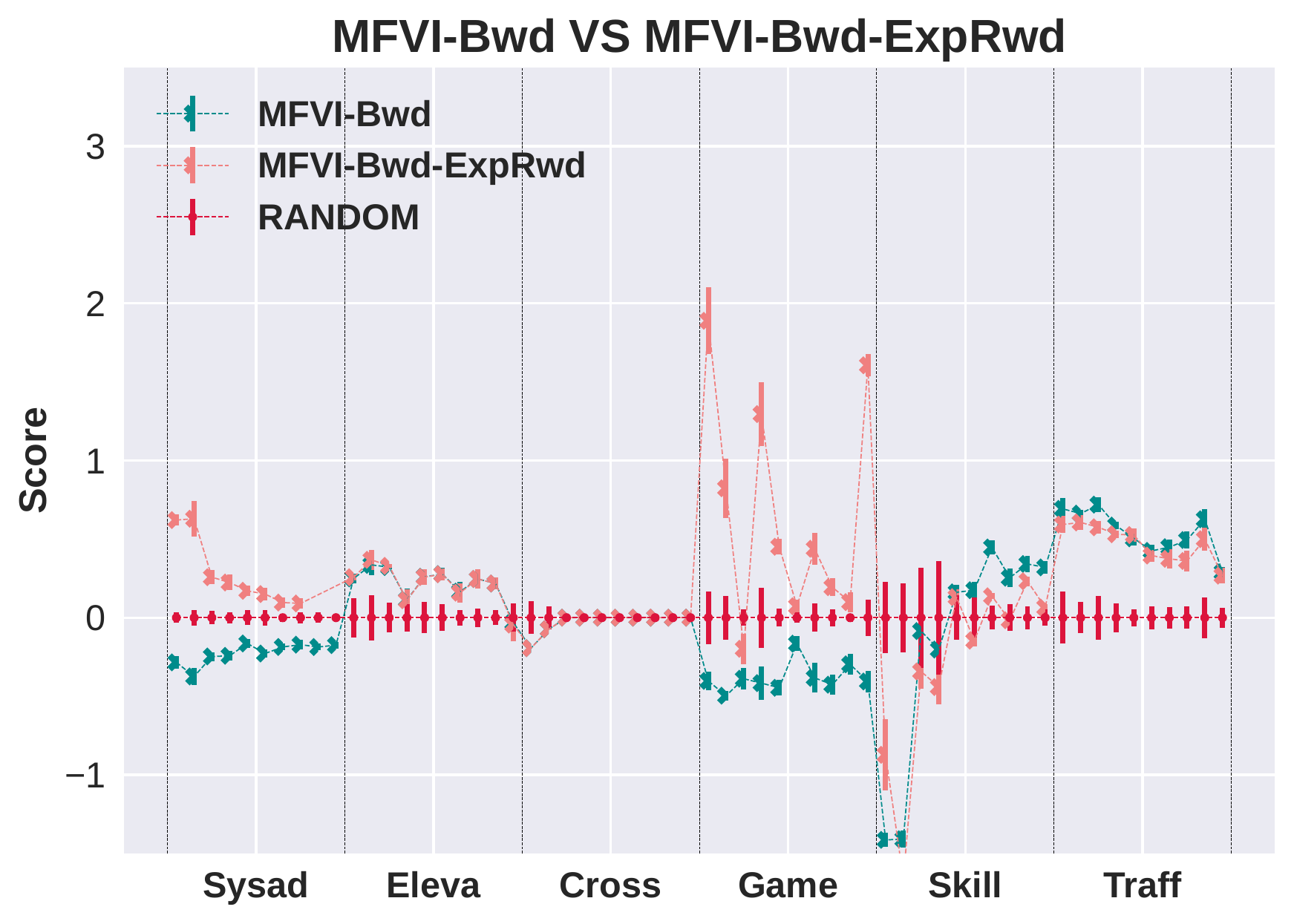}
	\end{subfigure}
	\begin{subfigure}[t]{0.32\linewidth}
		\centering
		\includegraphics[width=\linewidth, trim=7 7 7 7, clip]{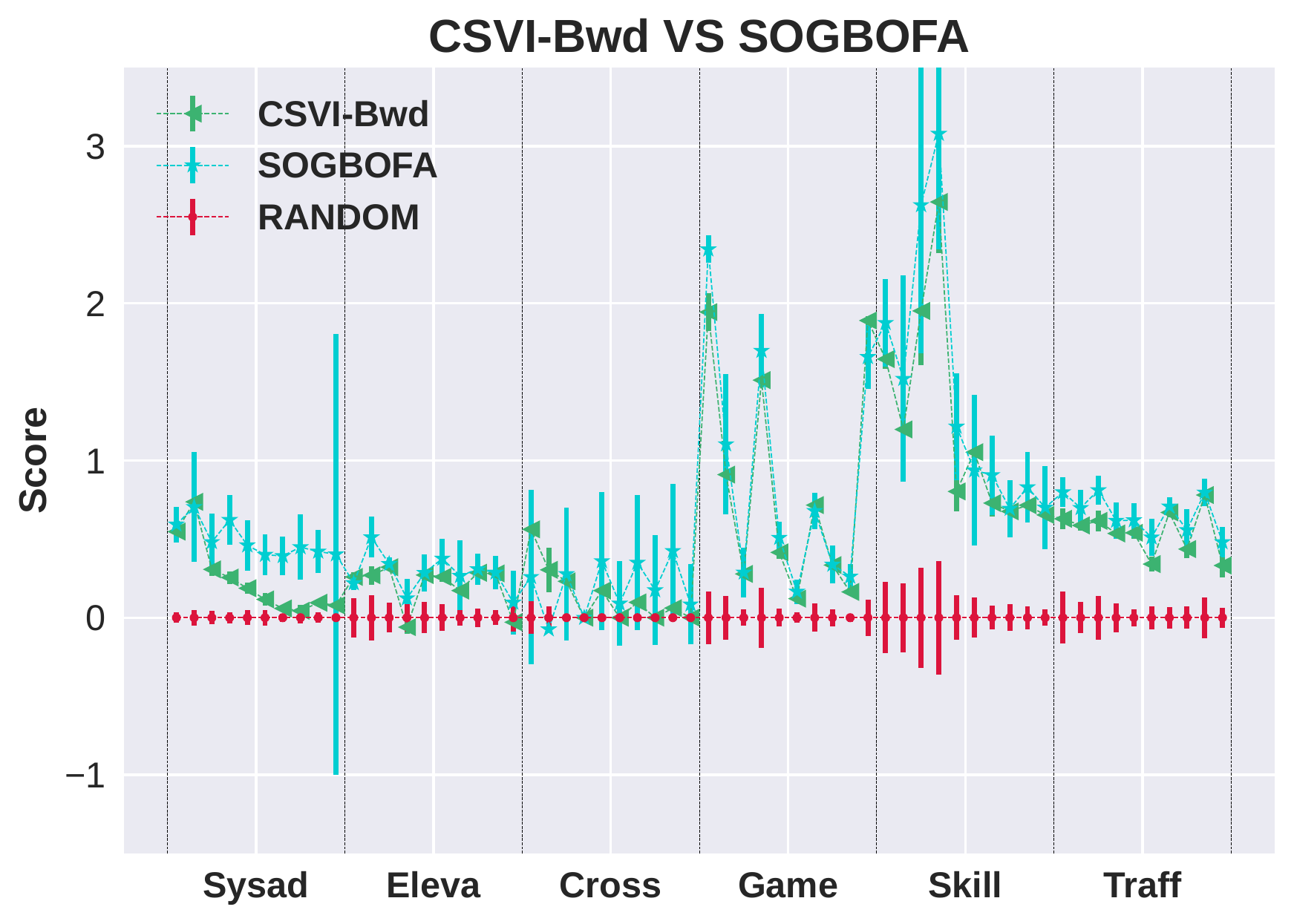}
	\end{subfigure}
	\begin{subfigure}[t]{0.32\linewidth}
		\centering
		\includegraphics[width=\linewidth, trim=7 7 7 7, clip]{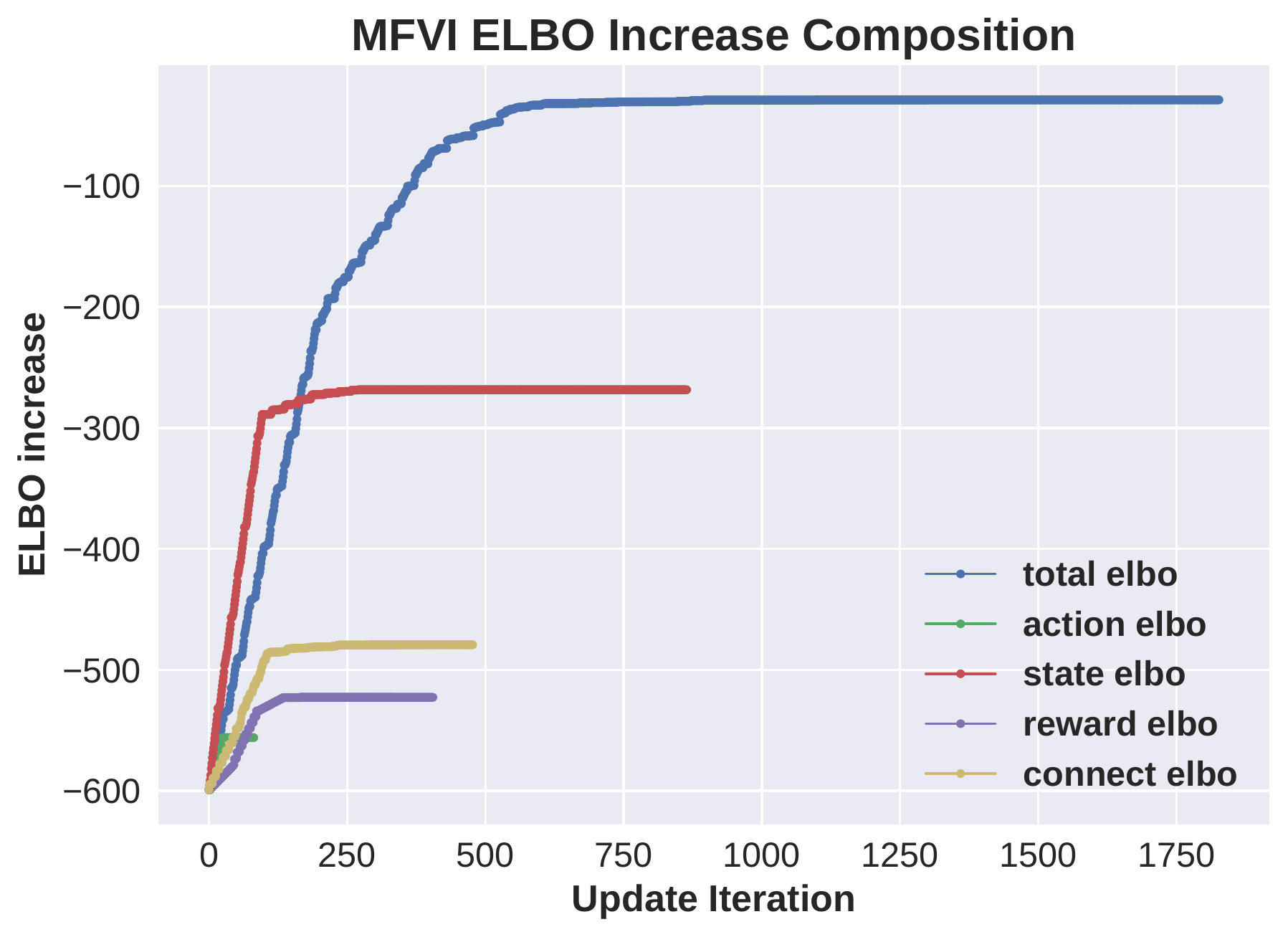}
	\end{subfigure}
\caption{
 First and second columns: algorithm comparisons on 60 problem instances, averaged over 12 simulations on each instance.
  Third column: comparing MFVI variants with and without state updates and the contributions of variable groups to the increase in the ELBO (in Skill teaching, instance 1, step 2 of execution).
}
\label{fig:Alg-Comp}
\end{figure*}
This section presents an experimental evaluation of the algorithms.
The code for regenerating all the results is available on \href{https://github.com/Zhennan-Wu/AISPFS}{Github}\footnote{https://github.com/Zhennan-Wu/AISPFS}. Our goal in this paper is to understand the {\em quality of decisions} provided by different approximate inference 
schemes, ignoring implementation details. 
Therefore, 
during the experiments we do not limit run time 
but instead allow the algorithms to converge, within bounds given below, before proposing a decision. We chose 6 problem domains from the ICAPS 2011 International Probabilistic Planning Competition to conduct our experiments. Each domain has 10 instances with factorized structure, horizon of 40 and discount factor of 1, and instances differ by the number of state and action variables.
For our experiments we use the SPUDD \citep{hoey2013spudd} translation of the original RDDL \citep{Sanner2010relational} specification, which compiles away action factoring.
This simplifies the implementation because it removes the need to reconcile action constraints with factoring. 
To 
control our overall experimental time we use online planning with receding horizon control, where we set the search horizon to be the minimum value between 9 and the remaining time steps.

Algorithmic parameters for MFVI:
we perform at most 100 Variational updates and stop early if the infinity norm of the difference between consecutive approximation distributions is less than 0.1. We perform 3 outer iterations, i.e., policy updates for the forward version.

Algorithmic parameters for BP variants:
We use SOGBOFA \citep{cui2019stochastic}\footnote{https://github.com/hcui01/SOGBOFA} 
as forward Loopy BP,
fixing search depth to 9, and limiting the number of gradient updates to 500.
We note that SOGBOFA has outperformed other planners, including search based planners, 
in IPPC 2018 problems and is a state of the art baseline for the evaluation. 
For the backward algorithm, our implementation is based on \citet{zhou2022pgmax} with parallel message update and a bound of 100 iterations with no damping ($\beta=0$).
 
 Normalized mean $\pm$ one standard deviation 
 of the cumulative reward over 12 simulations
 are shown in all the plots. 
 Denote the mean value and standard deviation of the cumulative reward of algorithm $a$ on instance $i$ to be $\bar{r}_i(a)$, $\sigma_i(a)$, respectively. 
To facilitate comparisons across domains we report scores normalized relative to the random policy.
Specifically, for algorithm $a$ on instance $i$, $\mbox{score-mean}_i(a) = \frac{|\bar{r}_i(a) - \bar{r}_i(RANDOM)|}{|\bar{r}_i(RANDOM)|}$ and $\mbox{score-std}_i(a) = \frac{\sigma_i(a)}{|\bar{r}_i(RANDOM)|}$
where the random algorithm has score 0 and higher scores indicates better performance. 
For reference, the raw results are given in the \reftosupp.

\paragraph{Comparison of Algorithms}
Results are shown in the left column of Figure~\ref{fig:Alg-Comp}.
The top plot shows that search direction is important for BP: 
the forward algorithm (SOGBOFA) outperforms the backward algorithm.\footnote{ 
While our focus is on the quality of approximation it is worth noting that 
 \citet{cui2018stochastic} have shown that with a directed model (equivalent to the Forward Framework with no downstream evidence as in our case), LBP converges in one iteration. Thus the forward algorithm is also faster. 
}
In contrast, the second plot shows that for MFVI, there is no significant difference between the forward and backward variants.
This is an interesting result because, as shown above, the forward algorithms mimic Policy Iteration and they provide a tighter approximation. 
%
The third plot compares 
MFVI to BP showing that MFVI has poor performance in some problems and forward BP dominates in all problems. 
Finally, 
we can show 
(see \reftosupp)
that the exponentiated variant of MFVI can be captured in our framework by conditioning on all reward and cumulative reward variables. 
The bottom plot compares 
this variant to standard MFVI. We see that the performance improves in two domains but 
the exponentiated variant
is still dominated by forward BP. 


\paragraph{Exploring the performance of MFVI} 
We believe that the main reason for the failure of MFVI is due to interaction between the flexibility that the mean field approximation allows with many state variables, and the sensitivity to ordering of updates due to local optima. 
To explore this we performed several additional experiments. In the first we introduce a new variant algorithm, MFVI-NoS, which does not update the marginal distribution over state variables, i.e. keeps them at the initialized value of $0.5$. Results for two domains are shown in the top half of the third column of Figure~\ref{fig:Alg-Comp}.
We see that while the NoS variant restricts the algorithm it improves the performance in these domains (this does not happen in all domains). Another view of this phenomenon is given by the relative contribution of each group of variables to the increase in the ELBO during updates of variational parameters. 
The bottom half of the third column of Figure~\ref{fig:Alg-Comp}
visualizes this for the MFVI and MFVI-NoS variants in one problem.
We see that for MFVI the largest increase in ELBO is contributed by adjusting state variables
and the NoS variant increases the share of other variables.
We further explore this in the 
full paper
using an artificial problem, showing that in this case
limiting the flexibility of MFVI can 
lead to better posterior, that MFVI is sensitive to a choice of which subset of state variables is updated, and in addition to the order of updates.


\section{CSVI}\label{sec:csvi}

Motivated by the analysis above, we propose a new algorithm for variational inference in planning. Instead of treating all the latent nodes in the DBN in the same manner and
computing approximate distributions over all these variables, the algorithm focuses on the action variables and effectively 
marginalize out other terms to achieve a tighter {ELBO}. This type of approach is known as collapsed variational inference, which has been shown to be effective in models where the marginalization can be done analytically (e.g., \citet{teh2006collapsed}) but for planning one has to resolve additional computational challenges as we show below. Specifically we propose to use the following provably tighter ELBO
\begin{equation}
     \quad \log p_{\theta}(c_T = 1) 
    = \log \mathbb{E}_{p_{\theta(A)}}[p(c_T = 1,A)]
     \geq
    \mathbb{E}_{q_{\phi}}[\log \frac{p_{\theta}(c_T = 1,A)}{q_{\phi}(A)}].
\end{equation}
Here we have
the same factorized transitions and policy distribution. However, we do not compute approximation distributions over state, reward, and cumulative reward variables. 
With mean field, the standard solution \citep{bishop2006pattern}
yields the update equation
\begin{align}
    \log q_{\phi}(a_t^l) \propto \mathbb{E}_{q_{\phi} \backslash a_t^l (A) } \log p_{\theta}(c_T = 1, A) =  \mathbb{E}_{q_{\phi} \backslash a_t^l(A) }\log g_{\theta}(A)
\label{eq:csviupdatea}
\end{align}
\begin{align}
    \intertext{where}
    g_{\theta}(A) = \mathbb{E}_{S, R, C_{\setminus T}} [p_{\theta}(A, S, R, C, c_T = 1)].
\end{align}
The tighter approximation appears to yield an infeasible update, because $A$ is entangled in $g()$ and we must perform an explicit marginalization in $g()$ for each update.

We next show how the update equation can be approximated via sampling. The key is to first extract $p_{\theta}(A)$ from the expectation. We therefore have:
\begin{align}
    \log g_{\theta}(A) = \log  p_{\theta}(A) + \log \mathbb{E}_{S, R, C_{\setminus T}} [p_{\theta}(S, R, C, c_T = 1 | A)].
\end{align}
Recall that $p_{\theta}(A)$ is a product of independent terms.
This implies that the first part can be substituted with $\log p_{\theta}(a^l_h)$ since all other terms are constants w.r.t the variable of interest in (\ref{eq:csviupdatea}) and they will vanish in the normalized update of $q_{\phi}(a_t^l)$. 
The second part is conditioned on $A$ and does not include $p(A)$ terms. Its expectation can be estimated through sampling.
In particular, sampling can be intuitively done as follows: keeping $a^l_t$ fixed, sample the action sequence from approximate distribution $q_{\phi\backslash a^l_t}(A)$. Then complement this by sampling values for $s_t$, $r_t$, $c_t$ nodes, including $c_T$. The resulting values for $c_T$ are generated from the correct distribution and the average over $c_T$ gives an estimate of the expectation. 
Since we are using sampling and averaging inside the logarithm this yields biased estimates for updates,
but this type of biased estimates has been shown to work in other cases in machine learning 
(e.g., \citep{wei2021direct})
and it can be mitigated by taking sufficient samples. 
It is interesting to note from the above update that the policy distribution serves as a weight 
bias in the action update procedure. Algorithm~\ref{alg:CSVI} summarizes the update procedure.

\begin{algorithm}
	\caption{Collapsed State Variational Inference} 
	\begin{algorithmic}[1]
		\For {$t=1,2,\ldots, T$}
		    \For {$l = 1, 2, \ldots, N$}
			    \For {value of action variable $l$ at time $t$ fixed to be $0, 1$}
			        \For {action sequence sample index $i = 1, \ldots, M_1$}
				        \State Sample action sequence $A = a_{1},\ldots,a_{T}$ from $q_{\phi}$
				        \For {trajectory sample $= 1, \ldots, M_2$}
				            \State Sample and record cumulative reward variable $c_T$ from $g_{\theta}(A)$
				        \EndFor
				        \State Estimate $\hat{p}_i = \#(c_T = 1)/M_2$
				    \EndFor
				    \State Calculate $\log q_{\phi}(a^l_t) \propto \log p_{\theta}(a^l_t) + \sum_i (\log \hat{p}_i)/M_1 $
			    \EndFor
            \State Update $q_{\phi}(a_t^l)$ by calculating the normalizing factor
			\EndFor            
		\EndFor
	\end{algorithmic} 
\label{alg:CSVI}
\end{algorithm}

\paragraph{Performance of CSVI} 
For CSVI our implementation uses the same parameters as in MFVI except that we make at most 10 variational updates.
The sample sizes are set to $M_1=20$ and $M_2=50$. 
Results are shown in the middle column of Figure~\ref{fig:Alg-Comp}.
Considering the plots from top to bottom we observe that 
there is no significant difference between forward and backward variants of CSVI and that 
CSVI is significantly better than MFVI.
The third plot shows that
the exponentiated reward variant does not improve the performance of CSVI. 
This suggests that the improvement over exponential variant for MFVI is due to stabilizing the optimization rather than presenting a better objective. 
The fourth plots shows that 
the performance of CSVI is competitive with forward BP and therefore CSVI provides state of the art performance in stochastic planning. 


\section{Conclusion}
In this paper we provide a unified scheme that categorizes many previous approaches along two dimensions, using either forward or backward reasoning and choosing an approximation scheme. 
Specifically, we focus on belief propagation and mean field variational inference as the approximation choices.
In this context, we illustrate the advantage of Forward Loopy BP as providing the best performance. 
Algorithms based on MFVI perform poorly in some domains. They are improved by exponential reward weighting but not sufficiently so. An experimental analysis points to sensitivity of the optimization as a source for this failure. 
Motivated by this analysis we propose a novel algorithm, Collapsed State Variational Inference, which provides a tighter variational approximation, and while being computationally demanding it performs competitively with Forward Loopy BP.
The results highlight that
while BP has been less in focus in recent years, it provides a strong baseline for stochastic planning.
It also shows the importance of focusing variational approximations on variables of interest as done in CSVI and the potential for developing strong variational algorithms for planning.  %
%
These observations suggest interesting directions for future work including developing efficient variants of CSVI, using amortized variational inference in planning to improve CSVI, alternative schemes to capture the posterior distributions in VI, and developing tighter approximations and optimization algorithms through BP methods.

\begin{acknowledgements}
This work was supported by NSF under grant IIS-1906694 and grant IIS-2002393. Some of the experiments in this paper were run on the Big Red 3 computing system at Indiana University, supported in part by Lilly Endowment,Inc., through its support for the Indiana University Pervasive Technology Institute.
\end{acknowledgements}
\bibliography{uai2022-template}

\clearpage
\appendix
\onecolumn
\section{Equivalence of Different Reward Formulations}
In the MDP framework, we are trying to maximize the cumulative reward. We first show that this is captured by the sum of binary reward variables.
\begin{align}
    U(\pi_{\theta}) &= \sum_{\substack{a_0 \\ s_t \in \mathcal{S}\\ a_{t} \in \mathcal{A} \\ t\in 1, \cdots, T}}\Big[\prod_{t=1}^Tp(s_t|s_{t-1}, a_{t-1})p_{\theta}(a_{t-1}|s_{t-1})\Big] \big(\sum_{h=1}^T R(s_h,a_h)\big) \notag \\
    &= \sum_{h=1}^T\sum_{\substack{a_0 \\s_t \in \mathcal{S} \notag \\
    a_t \in \mathcal{A}\\ t\in 1, \cdots, h}}\prod_{t=1}^{h}p(s_t|s_{t-1}, a_{t-1})p_{\theta}(a_{t-1}|s_{t-1})R(s_h,a_h)\\
    &= \sum_{h=1}^T \mathbb{E_{\theta}}(R(s_h,a_h)) \\
    &\propto \sum_{h=1}^T \mathbb{E_{\theta}}(r_h).
\end{align}

\subsection{Cumulative Binary Reward Over time}
    Under the intermediate DBN setting, we need to calculate the expectation of the total reward $\tilde{R}$. We have:
    \begin{align}
        \mathbb{E}_{\theta}(\tilde{R}) &= p_{\theta}(\tilde{R} = 1) \notag\\ 
        &= \sum_{\substack{a_0 \\s_t \in \mathcal{S} \\ a_t \in \mathcal{A} \\ r_t \in \{0, 1\} \\ t = 1, \cdots T}}\Big[\big[\prod_{t=1}^{T}p(s_t|s_{t-1}, a_{t-1})p_{\theta}(a_{t-1}|s_{t-1})p(r_t|s_{t-1}, a_{t-1})\big]p(\tilde{R} = 1|r_1, \cdots, r_T)\Big] \notag \\
        &\propto\sum_{\substack{a_0 \\s_t \in \mathcal{S} \\ a_t \in \mathcal{A} \\ r_t \in \{0, 1\} \\ t = 1, \cdots T}}\Big[\big[\prod_{t=1}^{T}p(s_t|s_{t-1}, a_{t-1})p_{\theta}(a_{t-1}|s_{t-1})p(r_t|s_{t-1}, a_{t-1})\big]\frac{\sum_{j=1}^T r_j}{T}\Big] \notag \\
        &= \sum_{j=1}^T\sum_{\substack{a_0 \\s_t \in \mathcal{S} \\ a_t \in \mathcal{A} \\ r_t \in \{0, 1\} \\ t = 1, \cdots T}}\Big[\big[\prod_{t=1}^{T}p(s_t|s_{t-1}, a_{t-1})p_{\theta}(a_{t-1}|s_{t-1})p(r_t|s_{t-1}, a_{t-1})\big]\frac{r_j}{T}\Big] \notag
    \end{align}
    \begin{align}
        &= \frac{1}{T}\sum_{\substack{a_0 \\s_t \in \mathcal{S} \\ a_t \in \mathcal{A} \\ r_t \in \{0, 1\} \\ t = 1, \cdots T}}\Big[\big[\prod_{\substack{t=1 \\ t \neq j}}^{T}p(s_t|s_{t-1}, a_{t-1})p_{\theta}(a_{t-1}|s_{t-1})p(r_t|s_{t-1}, a_{t-1})\big] \notag\\
        & \qquad p(s_j|s_{j-1}, a_{j-1})p_{\theta}(a_{j-1}|s_{j-1})p(r_j=1|s_{j-1}, a_{j-1})\Big] \\
        & \mbox{Marginalize out all terms with index $i > j$} \notag\\
        &= \frac{1}{T}\sum_{j=1}^{T}\sum_{\substack{a_0 \\s_t \in \mathcal{S} \\ a_t \in \mathcal{A} \\ r_t \in \{0, 1\} \\ t = 1, \cdots T}}\Big[\big[\prod_{t=1}^{j-1}p(s_t|s_{t-1}, a_{t-1})p_{\theta}(a_{t-1}|s_{t-1})p(r_t|s_{t-1}, a_{t-1})\big]\notag\\
        & \qquad p(s_j|s_{j-1}, a_{j-1})p_{\theta}(a_{j-1}|s_{j-1})p(r_j = 1|s_{j-1}, a_{j-1})\Big] \\
        &\mbox{Marginalize out all the reward terms with index $i \neq j$} \notag\\
        &= \frac{1}{T}\sum_{j=1}^T\sum_{\substack{a_0 \\s_t \in \mathcal{S} \\ a_t \in \mathcal{A} \\ t = 1, \cdots j}}\Big[\big[\prod_{t=1}^j p(s_t|s_{t-1}, a_{t-1})p_{\theta}(a_{t-1}|s_{t-1})\big]p(r_j = 1|s_{j-1}, a_{j-1})\Big] \notag\\
        &= \frac{1}{T}\sum_{j=1}^T \mathbb{E}_{\theta}(r_j)\\
        &\propto \sum_{h=1}^T \mathbb{E}_{\theta}(r_h).
    \end{align}

Therefore the intermediate DBN with $\tilde{R}$ is also captured by the sum of binary reward variables.

For our final proposed DBN, we have the expectation:
    \begin{align}
        \mathbb{E}_{\theta}(c_T = 1) &=  p_{\theta}(c_T = 1) \notag\\
        &= \sum_{\substack{a_0 \\s_t \in \mathcal{S} \\ a_t \in \mathcal{A} \\ r_t, c_t \in \{0, 1\} \\ t = 1, \cdots T-1}} \Big[\prod_{t=1}^{T-1}p(s_t|s_{t-1}, a_{t-1})p_{\theta}(a_{t-1}|s_{t-1})p(r_t|s_{t-1}, a_{t-1})p(c_t|c_{t-1}, r_t) \notag\\
        &\qquad p(s_T|s_{T-1}, a_{T-1})p_{\theta}(a_{T-1}|s_{T-1})p(r_T|s_{T-1}, a_{T-1})p(c_T = 1|c_{T-1}, r_T)\Big] \notag
    \end{align}
    \begin{align}
        &= \sum_{\substack{a_0 \\s_t \in \mathcal{S} \\ a_t \in \mathcal{A} \\ r_t, c_t \in \{0, 1\} \\ t = 1, \cdots T-1}}\Big[\prod_{t=1}^{T-1}p(s_t|s_{t-1}, a_{t-1})p_{\theta}(a_{t-1}|s_{t-1})p(r_t|s_{t-1}, a_{t-1})p(c_t|c_{t-1}, r_t) \notag\\
        &\qquad p(s_T|s_{T-1}, a_{T-1})p_{\theta}(a_{T-1}|s_{T-1})p(r_T|s_{T-1}, a_{T-1})\frac{(T-1)c_{T-1} + r_T}{T}\Big] \notag \\
        &= \sum_{\substack{a_0 \\s_t \in \mathcal{S} \\ a_t \in \mathcal{A} \\ r_t, c_t \in \{0, 1\} \\ t = 1, \cdots T-1}}\Big[\prod_{t=1}^{T-1}p(s_t|s_{t-1}, a_{t-1})p_{\theta}(a_{t-1}|s_{t-1})p(r_t|s_{t-1}, a_{t-1})p(c_t|c_{t-1}, r_t)\notag\\ 
        &\qquad  p(s_T|s_{T-1}, a_{T-1})p_{\theta}(a_{T-1}|s_{T-1})p(r_T|s_{T-1}, a_{T-1})\frac{T-1}{T}c_{T-1}\Big] \notag\\
        & \quad + \sum_{\substack{a_0 \\s_t \in \mathcal{S} \\ a_t \in \mathcal{A} \\ r_t, c_t \in \{0, 1\} \\ t = 1, \cdots T-1}}\Big[\prod_{t=1}^{T-1}p(s_t|s_{t-1}, a_{t-1})p_{\theta}(a_{t-1}|s_{t-1})p(r_t|s_{t-1}, a_{t-1})p(c_t|c_{t-1}, r_t) \notag\\
        &\qquad p(s_T|s_{T-1}, a_{T-1})p_{\theta}(a_{T-1}|s_{T-1})p(r_T|s_{T-1}, a_{T-1})\frac{r_T}{T}\Big] \\
        &\mbox{Marginalizing out $s_T$, $a_{T-1}$ and $r_T$ in the first part of the equation} \notag\\
        & \triangleq \frac{T-1}{T} A_{T-1} + \frac{1}{T} B_T 
    \end{align}
    where
    \begin{align}
        A_{T-1} &= \sum_{\substack{a_0 \\s_t \in \mathcal{S} \\ a_t \in \mathcal{A} \\ r_t, c_t \in \{0, 1\} \\ t = 1, \cdots T-2}}\Big[\prod_{t=1}^{T-2}p(s_t|s_{t-1}, a_{t-1})p_{\theta}(a_{t-1}|s_{t-1})p(r_t|s_{t-1}, a_{t-1})p(c_t|c_{t-1}, r_t) \notag\\
        &\qquad p(s_{T-1}|s_{T-2}, a_{T-2})p(r_{T-1}|s_{T-2}, a_{T-2})p(c_{T-1} = 1|c_{T-2}, r_{T-1})\Big]\\
        B_T &= \sum_{\substack{a_0 \\s_t \in \mathcal{S} \\ a_t \in \mathcal{A} \\ r_t, c_t \in \{0, 1\} \\ t = 1, \cdots T-1}}\Big[\prod_{t=1}^{T-1}p(s_t|s_{t-1}, a_{t-1})p_{\theta}(a_{t-1}|s_{t-1})p(r_t|s_{t-1}, a_{t-1})p(c_t|c_{t-1}, r_t) \notag\\
        & \qquad p(s_T|s_{T-1}, a_{T-1})p(a_{T-1}|s_{T-1})p(r_T|s_{T-1}, a_{T-1})r_T\Big].
    \end{align}

Notice that 
    \begin{equation}
        p_{\theta}(c_T = 1) = A_T.
    \end{equation}

    For part $B_T$, since $r_T$ is binary
    \begin{align}
        & B_T = \frac{1}{T}\sum_{\substack{a_0 \\s_t \in \mathcal{S} \\ a_t \in \mathcal{A} \\ r_t, c_t \in \{0, 1\} \\ t = 1, \cdots T-1}}\Big[\prod_{t=1}^{T-1}p(s_t|s_{t-1}, a_{t-1})p_{\theta}(a_t|s_t)p(r_t|s_{t-1}, a_{t-1})p(c_t|c_{t-1}, r_t) \notag\\
        &\qquad p(s_T|s_{T-1}, a_{T-1})p(a_{T-1}|s_{T-1})p(r_T = 1|s_{T-1}, a_{T-1})\Big]\\
        & \mbox{Marginalize out $c_t$, $r_t$ for $t = 1, \cdots, T-1$} \notag\\
        &= \frac{1}{T}\sum_{\substack{a_0 \\s_t \in \mathcal{S} \\ a_t \in \mathcal{A} \\ r_t, c_t \in \{0, 1\} \\ t = 1, \cdots T-1}}\Big[\prod_{t=1}^Tp(s_t|s_{t-1}, a_{t-1})p(a_{T-1}|s_{T-1})p(r_T = 1|s_{T-1}, a_{T-1})\Big] \notag\\
        &= \frac{1}{T}\mathbb{E}_{\theta}(r_T).
    \end{align}

    Given these observations we have the recursive equation
    \begin{align}
        A_T &= \frac{T-1}{T}A_{T-1} + \frac{1}{T}B_T \notag\\
        &= \frac{T-1}{T}A_{T-1} + \frac{1}{T}\mathbb{E}_{\theta}(r_T) \notag\\
        &= \frac{T-1}{T}(\frac{T-2}{T-1}A_{T-2} + \frac{1}{T-1}\mathbb{E}_{\theta}(r_{T-1})) + \frac{1}{T} \mathbb{E}_{\theta}(r_T)  \notag\\
        &= \frac{T-2}{T}A_{T-2} + \frac{1}{T}(\mathbb{E}_{\theta}(r_T) + \mathbb{E}_{\theta}(r_{T-1})) \notag\\
        &= \cdots \notag\\
        &= \frac{2}{T}A_{2} + \frac{1}{T}\sum_{t=3}^T\mathbb{E}_{\theta}(r_t) \notag\\
        &\propto \sum_{h=1}^T \mathbb{E}_{\theta}(r_h).
    \end{align}
    i.e., the expectation of $\tilde{R}$ and $c_T$ are equivalent in two DBNs and they are both proportional to the expected cumulative reward of the original MDP problem.

\section{Accumulating reward from multiple nodes in the same time step}
In large factored state and action spaces, the rewards are typically specified as an addition function over small factors
that only depend on a small number of state and action variables given by some decision rules. 
The sum variable might have many parents and therefore we require an addition construction for the DBN.
Since this construction is done for each time step separately, in this section we simplify the notation and omit the subscript of time step~$t$.

The construction is similar to the accumulation of reward over time. Assume there are $K$ decision rules to determine the reward at a particular time step with some state and action. Given some order over the decision rules, we expand the DBN so that each decision rule corresponds to a binary partial reward node $pr_i$, ($i = 1, \cdots K$), with edge between the partial reward node and the dependent state and action nodes according to the decision rule. Then for each partial reward node $pr_i$, we create a binary collecting reward node $cr_i$ that connects to the partial reward node $pr_i$ and the collecting reward node $cr_{i-1}$ of the previous partial reward node. We also create an additional collecting reward node $cr_0$ which is set to $1$.

We then define the conditional distribution of $cr_i$ given $cr_{i-1}$, $pr_i$ to be
\begin{align}
    p(cr_{i} = 1|cr_{i-1}, pr_{i}) = \frac{(i-1)cr_{i-1} + pr_i}{i}
\end{align}
and the partial reward distribution $pr_i$ to be
\begin{align}
    p(pr_i|s, a) \propto \mbox{$i$th reward decision rule}.
\end{align}

We want to show that $p(cr_K = 1|s, a) \propto r(s, a)$ for every time step. 

\begin{align}
    p(cr_K = 1|s, a) &= \sum_{\substack{cr_i, pr_i, pr_K \\ i = 1, \cdots, K-1}} p(cr_K = 1|cr_{K-1}, pr_K)p(pr_K|s, a)\prod_{i=1}^{K-1} p(cr_i|cr_{i-1}, pr_i)p(pr_i|s, a) \notag \\
    &= \sum_{\substack{cr_i, pr_i, pr_K \\ i = 1, \cdots, K-1}} \frac{(K-1)cr_{K-1} + pr_K}{K}p(pr_K|s, a)\prod_{i=1}^{K-1} p(cr_i|cr_{i-1}, pr_i)p(pr_i|s, a) \notag \\
    &\mbox{Separate the formula above w.r.t. $cr_{K-1}$ and $pr_K$}  \notag \\
    &= \frac{K-1}{K} \mbox{part1} + \frac{1}{K}\mbox{part2} 
\end{align}
where
\begin{align}
\mbox{part1} &= \sum_{\substack{cr_i, pr_i, pr_{K-1} \\ i = 1, \cdots, K-2}}  
    \big[p(cr_{K-1} = 1|cr_{K-2}, pr_{K-1}) \prod_{i=1}^{K-2} p(cr_i|cr_{i-1}, pr_i)p(pr_i|s, a) = p(cr_{k-1} = 1|s, a)\big]
\end{align}
because the whole equation vanishes when $cr_{K-1} = 0$, and $pr_K$ got marginalized out. In addition,
\begin{align}
    \mbox{part2} &= p(pr_K=1|s, a)
\end{align}
because all other variables are marginalized out.

Thus we have 
\begin{align}
    p(cr_K = 1|s, a) &= \frac{K-1}{K}p(cr_{K-1} = 1|s, a) + \frac{1}{K}p(pr_K = 1|s, a) \notag\\
    &= \cdots \notag\\
    &= \frac{1}{K}\sum_{i=1}^Kp(pr_i = 1|s, a) \notag\\
    &\propto \sum_{i} i\mbox{th reward decision rule} \notag\\
    &\propto r(s, a).
\end{align}

\section{Exponentially Weighted Reward}

As discussed in the main paper, some prior work uses backward variational inference but does so with an exponential reward weighting. Here we show how this setting can be captured within our framework.
Recall that \citet{levine2018reinforcement} formulates the objective function as
\begin{align}
  D_{KL}(\hat{p}(\tau)||p(\tau|O_{1\cdots T})) 
\end{align}
where
\begin{align}
    p(\tau|O_{1\cdots T+1}) &= \Big[\prod_{t=1}^Tp(s_{t}|s_{t-1}, a_{t-1})\Big]\exp\Big(\sum_{t=0}^{T}r(s_t, a_t)\Big) \\
    \hat{p}(\tau) &=  \prod_{t=1}^Tp(s_{t}|s_{t-1}, a_{t-1})\pi(a_{t-1}|s_{t-1}).
\end{align}
Here $O(1, \cdots T+1)$ are indicator random variables denoting ``optimality'' in time $t = 1, \cdots T+1$ and the trajectory distribution is with an implicit uninformative policy.\par
In our formulation, $c_t$ represents the cumulative reward up to time $t$, and we have established that $\mathbb{E}(c_T)\propto \mathbb{E}(R)$. Recall in our graphical model, a complete trajectory distribution is 
\begin{align}
    p(\tau) &= \prod_{t=1}^Tp(s_t|s_{t-1}, a_{t-1})u(a_{t-1})p(r_{t}|s_{t-1}, a_{t-1})p(c_{t}|c_{t-1}, r_t) \notag\\
    &\quad u(a_T)p(r_{T+1}|s_T, a_T)p(c_{T+1}|c_T, r_{T+1}).
\end{align}
To recover the joint probability of the formulation from \citet{levine2018reinforcement}, we need the following steps: 
\begin{enumerate}
    \item Change reward distribution 
    to $p(r_t|s_{t-1}, a_{t-1}) \propto \exp(R(s_{t-1}, a_{t-1}))$.
    \item The trajectory distribution need to conditioned on $r_{1\cdots T+1}=1, c_{1\cdots T+1}=1$.
\end{enumerate}
Then in our formulation we have
\begin{align}
    p^*(\tau|r_{1\cdots T+1} = 1, c_{1\cdots T+1} = 1) &\propto \Big[\prod_{t=1}^Tp(s_{t}|s_{t-1}, a_{t-1})\Big]\exp\Big(\sum_{t=0}^{T}r(s_t, a_t)\Big)
\end{align}
which is the same as the optimal trajectory distribution above. Then the objective of \cite{levine2018reinforcement} can be seen to minimize
\begin{align}
    d_{KL}(q_{\phi}(\tau)|p^*(\tau|r_{1\cdots T+1} = 1, c_{1\cdots T+1} = 1))
\end{align}
which is captured in our framework with the backward VI by using additional observation variables.
Using the same methodology as above, in our framework both forward and backward variants of this MFVI variant can be implemented. 

\section{Full closed form update formula for MFVI}
For completeness we list the full update formulas of MFVI: 
\begin{align}
    \log q_{\phi}(s_{h}^{j}) & \propto \mathbb{E}_{q_{\phi}}^{\backslash s_{h}^{j}}\Big[\log p_{\theta}(S, A, R, C_{\backslash T}, c_T = 1)\Big]\notag\\
    & = \mathbb{E}_{q_{\phi}}^{\backslash s_{h}^{j}}\Big[\sum_{m=1}^M\log p(s_{h+1}^m|s_h, a_h, s_{h+1}^{G_{m}}) + \log p(s_h^{j}|s_{h-1}, a_{h-1}, s_h^{G_j})\notag\\
    &\qquad + \sum_{i : j \in G_i} \log p(s_h^i|s_{h-1}, a_{h-1}, s_h^{G_i}) + \log p(r_{h+1}|s_h, a_h)\Big]\\
    \notag \\
    \log q_{\phi}(a_h^l) & \propto \mathbb{E}_{q_{\phi}}^{\backslash a_h^l}\Big[\log p_{\theta}(S, A, R, C_{\backslash T}, c_T = 1)\Big]\notag \\
    &= \mathbb{E}_{q_{\phi}}^{\backslash a_h^l}\Big[\log p_{\theta}(a_h^l) + \log p(r_{h+1}|s_h, a_h) + \sum_{m=1}^M \log p(s_{h + 1}^m|s_h, a_h, s_{h+1}^{G_m})\Big]\\
    \notag \\
    \log q_{\phi}(r_h) &\propto \mathbb{E}_{q_{\phi}}^{\backslash r_h}\Big[\log p_{\theta}(S, A, R, C_{\backslash T}, c_T = 1)\Big]\notag \\
    & = \mathbb{E}_{q_{\phi}}^{\backslash r_h}\Big[\log p(r_h| s_{h-1}, a_{h-1}) + \log p(c_h|c_{h - 1}, r_h)\Big] \\
    \notag \\
    \log q_{\phi}(c_h) & \propto \mathbb{E}_{q_{\phi}}^{\backslash c_h}\Big[\log p_{\theta}(S, A, R, C_{\backslash T}, c_T = 1)\Big]\notag \\
    & = \mathbb{E}_{q_{\phi}}^{\backslash c_h}\Big[\log p(c_h|c_{h - 1}, r_h) + \log p(c_{h + 1}|c_h, r_{h + 1})\Big].
\end{align}

\section{Analysis of MFVI on a Demo Problem}

\begin{figure*}
	\centering
	\begin{subfigure}[t]{0.3\linewidth}
		\centering
		\includegraphics[width=\linewidth, trim=7 7 7 7, clip]{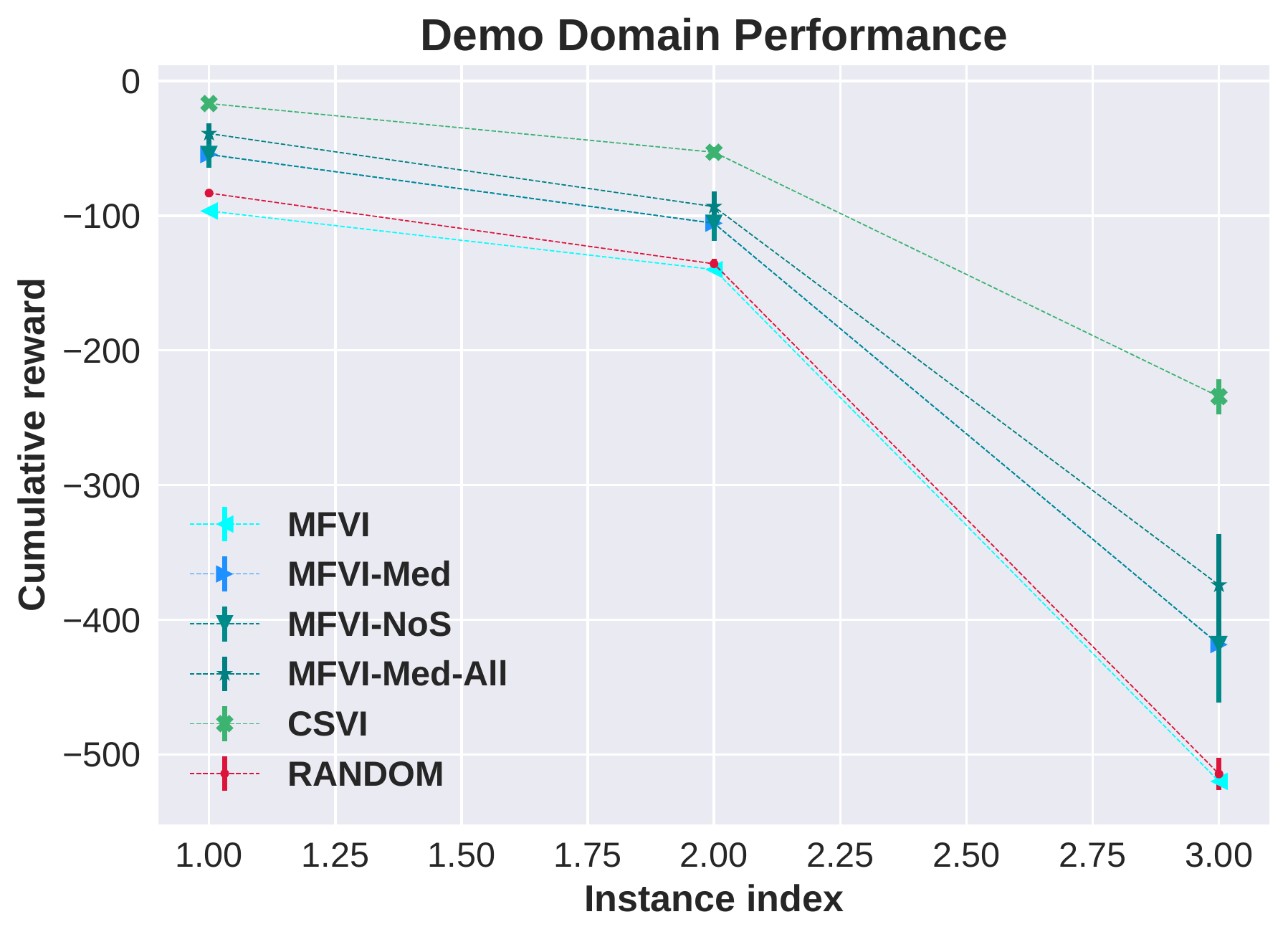}\label{fig:demo-performance}
	\end{subfigure}	
	\begin{subfigure}[t]{0.3\linewidth}
		\centering
		\includegraphics[width=\linewidth, trim=7 7 7 7, clip]{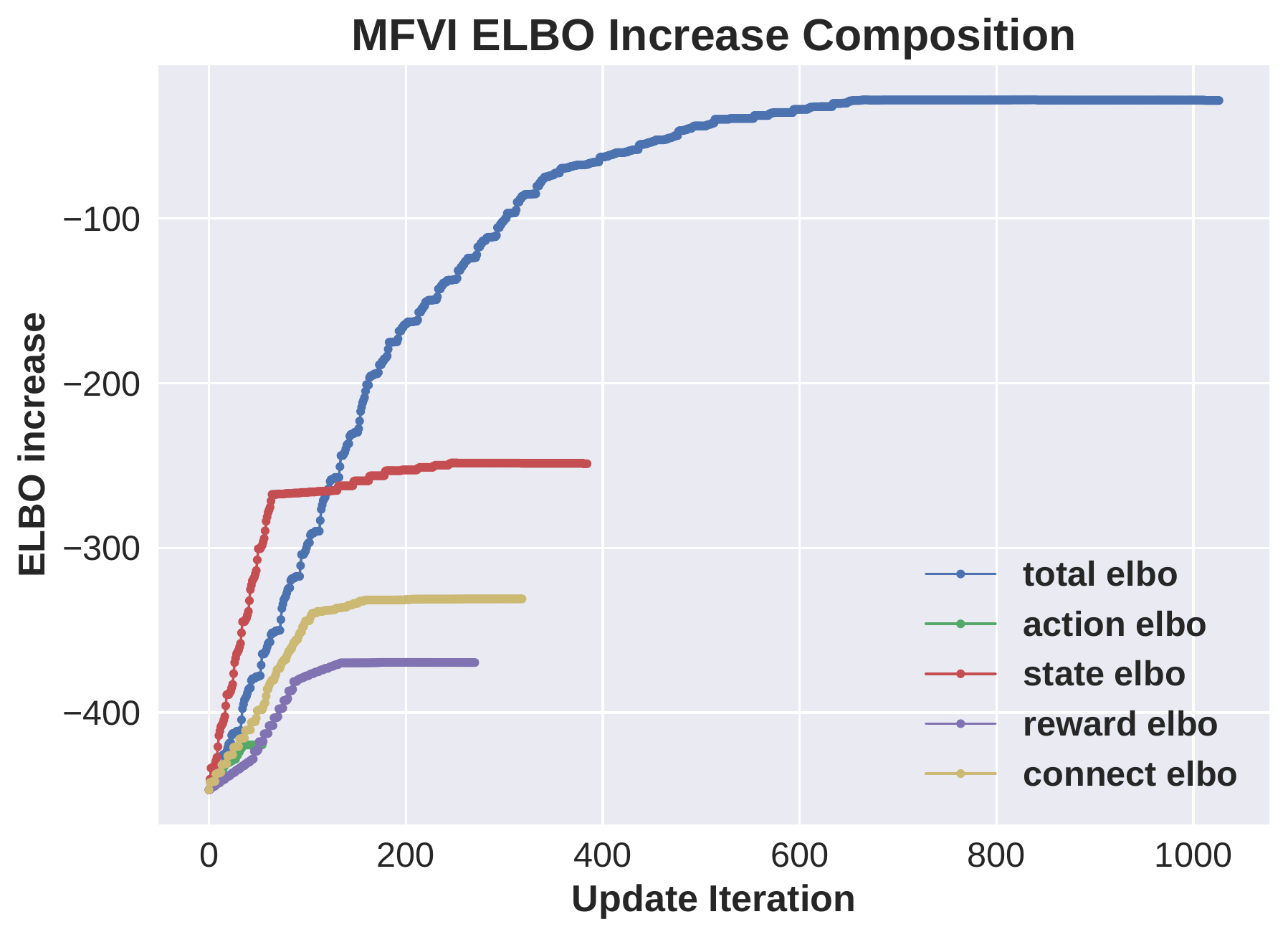}\label{fig:demo_all-elbo}
	\end{subfigure}	
	\begin{subfigure}[t]{0.3\linewidth}
		\centering
 		\includegraphics[width=\linewidth, trim=7 7 7 7, clip]{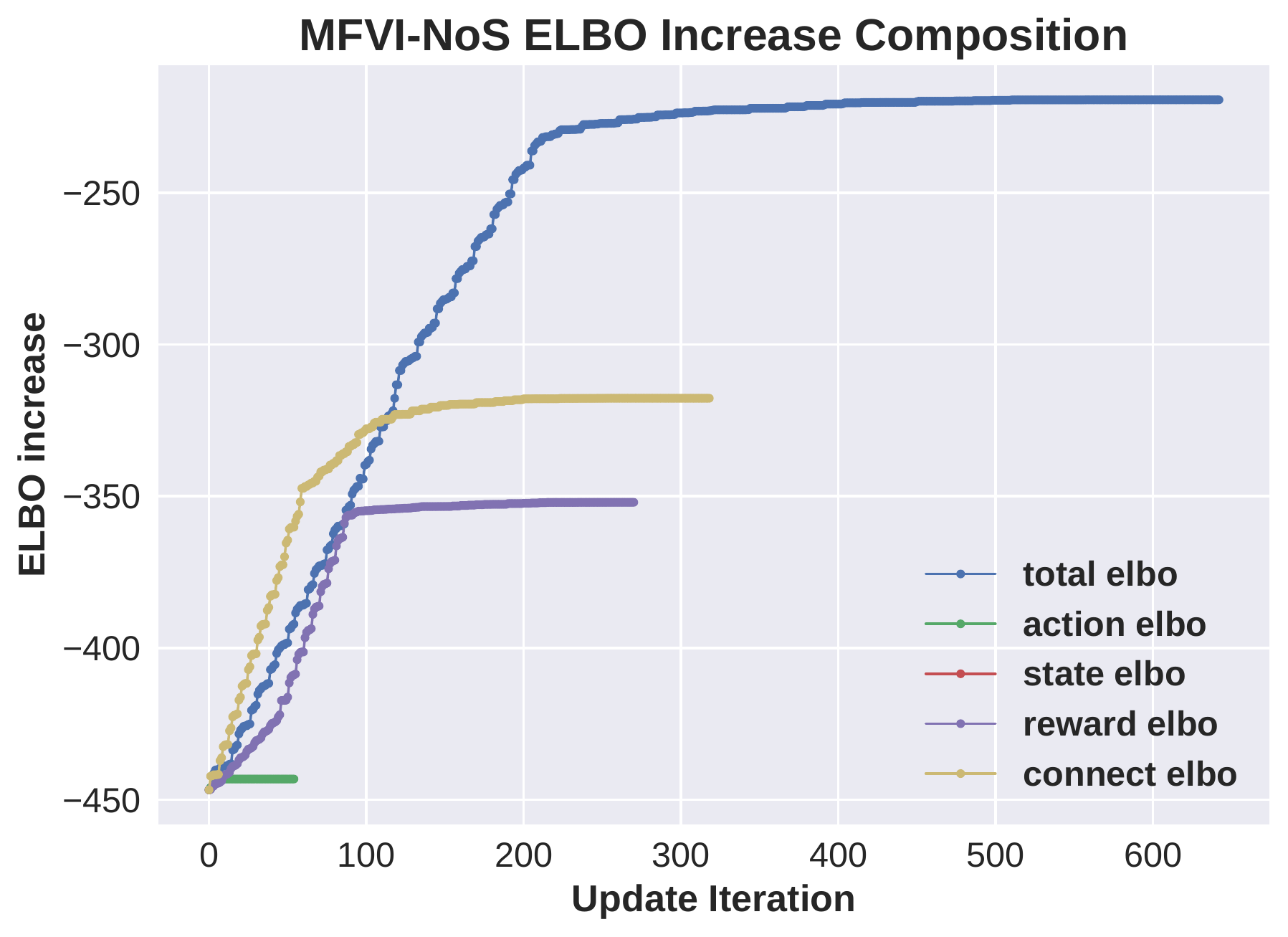}\label{fig:demo_noS-elbo}
	\end{subfigure}	
	\\
	\begin{subfigure}[t]{0.3\linewidth}
		\centering \ \ 
	\end{subfigure}			
	\begin{subfigure}[t]{0.3\linewidth}
		\centering
		\includegraphics[width=\linewidth, trim=7 7 7 7, clip]{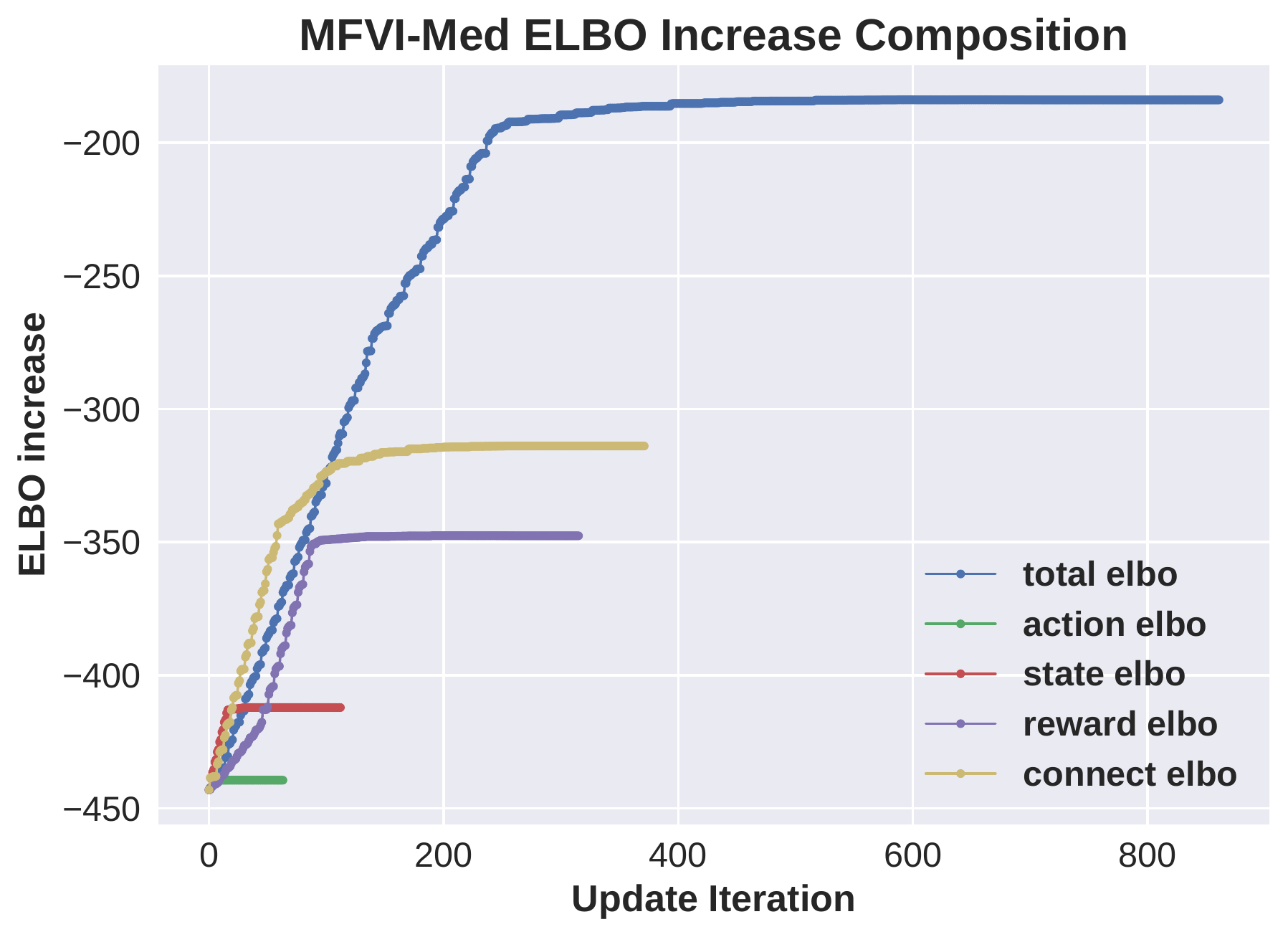}
	\end{subfigure}		
	\begin{subfigure}[t]{0.3\linewidth}
		\centering
		\includegraphics[width=\linewidth, trim=7 7 7 7, clip]{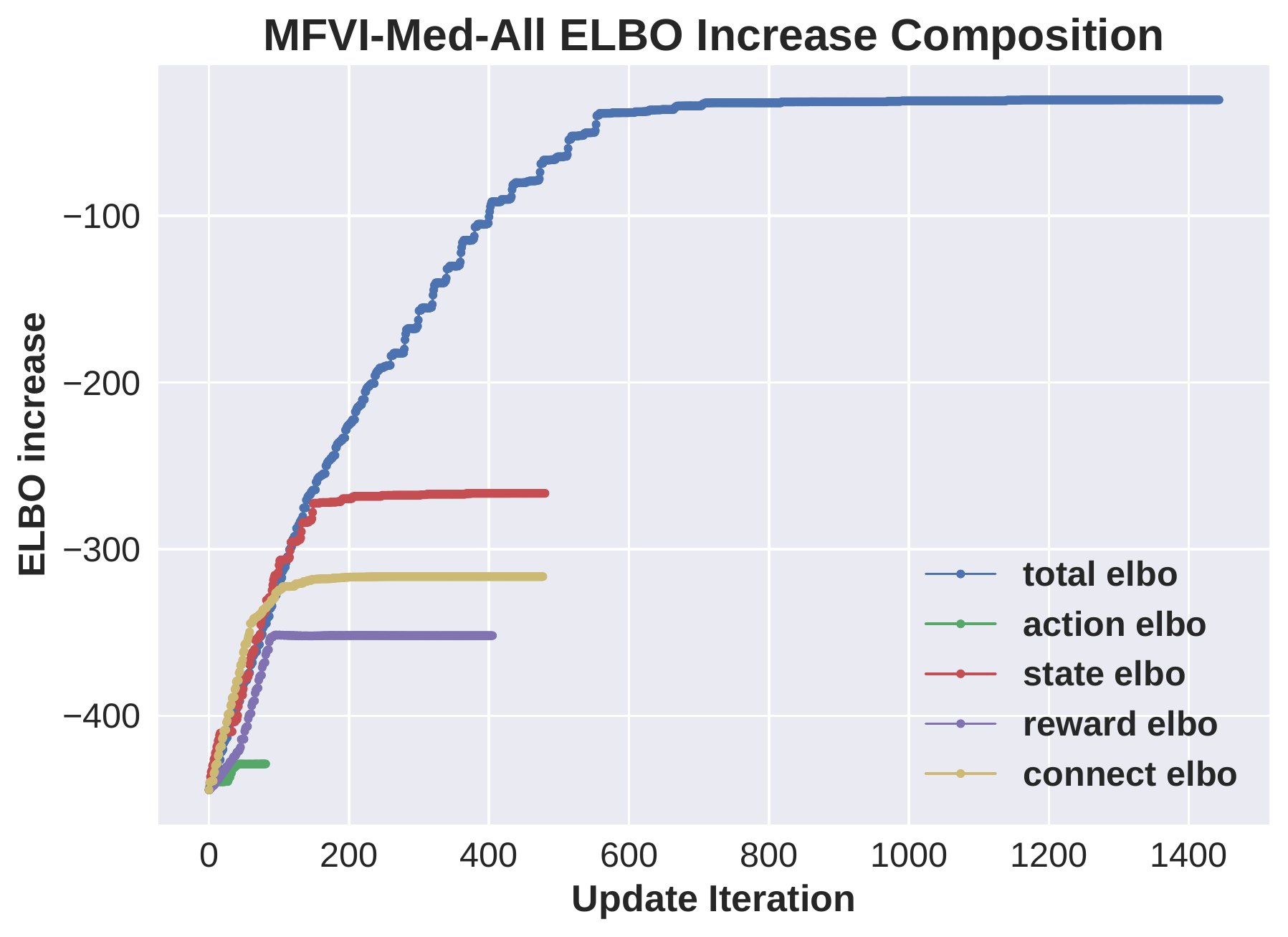}\label{fig:demo_med_all-elbo}
	\end{subfigure}
	\caption{Left: Performance of MFVI variants on the Cooking Problem. Other Plots: The increase in the value of the ELBO after updates of variables, shown for each group of variables separately.}\label{fig:demo}
\end{figure*}

In this section we further investigate the sensitivity of MFVI to updates of state variables. 
To achieve this, we design a new ``cooking'' domain, partly inspired by the structure of the Skill Teaching Domain, in the spirit of making inference intuitively clear. We have 2 dishes and three actions, ``cook dish1'', ``cook dish2'', ``do nothing''. 
By applying the ``cook'' action repeatedly each dish goes (with some probability) through 5 potential stages stating at ``not cooked'', and moving to ``not cooked and cooking'',
``cookMed'', ``cookMed and cooking'', ``cookWell'', ``Burned''. These stages are encoded by binary variables ``cookMed'', ``cookWell'', ``cooking''.
An additional state variable ``watching'' per dish,
which is equal to ``cooking'' adds flexibility while not affecting the dynamics.
The reward is given when ``cookMed'' and ``cookWell'' are true but not when ``burned''. 
Hence when conditioned on high cumulative reward we should expect to see "cookMed" initially set to true. 

We next explore algorithmic variants.
The NoS variant does not update state variables at all, as in the main paper.
Since ``cookMed'' is important
we create additional update schemes around it. ``MFVI-Med'' fixes the distribution of other state variables to be uniformly random and only updates ``cookMed'' state variables together with action, reward, cumulative reward variables. ``MFVI-Med-All'' performs an asynchronous update
stating by running `MFVI-Med''  to convergence and then following by running MFVI until it converges again.

Results are shown in Fig~\ref{fig:demo}. Consider first the plots that show increase in ELBO as a function of updates. We see that for MFVI the relative effect of state variables on the increase in ELBO is larger than all other variables.
Comparing this to the NoS variant we see that in that case reward and cumulative variables are more important and the improvement it provides is {\em potentially} due to removing the large changes in ELBO due to state variables. On the other hand, as shown in the performance plot on the left, the two new variants based on the ``cookMed'' variable, still improve the performance but also have a large gap in the effect on ELBO, so this does not provide a full explanation. The success of ``MFVI-Med-All'' shows that the flexibility is not the whole story, but that the algorithm is sensitive to the order of updates. 

As a final diagnostic, we print out the approximate posterior distribution of ``cookMed'', and ``cooking'' variables of different schemes starting from the state where all the variables are set to be 0. Ideally, these four variables should be all biased towards 1. We see that for VI-Med and VI-Med-All, their approximate posterior, though not fully accurate, provides useful information for action distribution update while MFVI provides approximate state posterior in the wrong direction.
This shows that MFVI can converge to uninformative local optima, which causes its poor performance. 
Overall we believe that the large number of state variables, their relative effect on the ELBO, and the sensitivity of the variational algorithm to order of updates are the cause of failure in some domains. 
\begin{table*}
    \centering
    \caption{Approximate posterior distribution of different MFVI variants at the initial state on the cooking problem.}\label{tab:demo-posterior}
    \begin{tabular}{ccccc}
      \toprule 
      \bfseries Variables & \bfseries VI-noS & \bfseries VI-Med & \bfseries VI-Med-All & \bfseries MFVI\\
      \midrule 
      t1-CookMed [d1, d2] & [0.5, 0.5] & [1.48e-3, 4.09e-6] & [3.37e-7, 5.47e-5] & [3.59e-5, 5.38e-5] \\
      t1-Cooking [d1, d2] & [4.95e-1, 4.95e-1] & [4.95e-1, 4.95e-1] & [\textbf{0.99}, 1.00e-12] & [5.00e-5, 5.00e-5] \\
      t2-CookMed [d1, d2] & [0.5, 0.5] & [\textbf{0.99}, 8.49e-4] & [\textbf{0.94}, 5.86e-5] & [7.57e-5, 5.90e-5] \\
      t2-Cooking [d1, d2] & [4.85e-1, 4.85e-1] & [4.67e-1, 4.67e-1] & [2.49e-1, 2.51e-1] & [5.00e-5, 5.00e-5] \\
      \bottomrule 
    \end{tabular}
\end{table*}

\section{Details Experimental Results in All Domains}
In this section we show the raw, un-normalized results separately for all domains (represented by the first five letters of the domain names in the main paper). We include results for a second implementation of forward Loopy BP which is discussed in the following subsection.
\begin{figure*}[ht]
	\centering
	\begin{subfigure}[t]{0.3\linewidth}
		\centering
		\includegraphics[width=\linewidth, trim=7 7 7 7, clip]{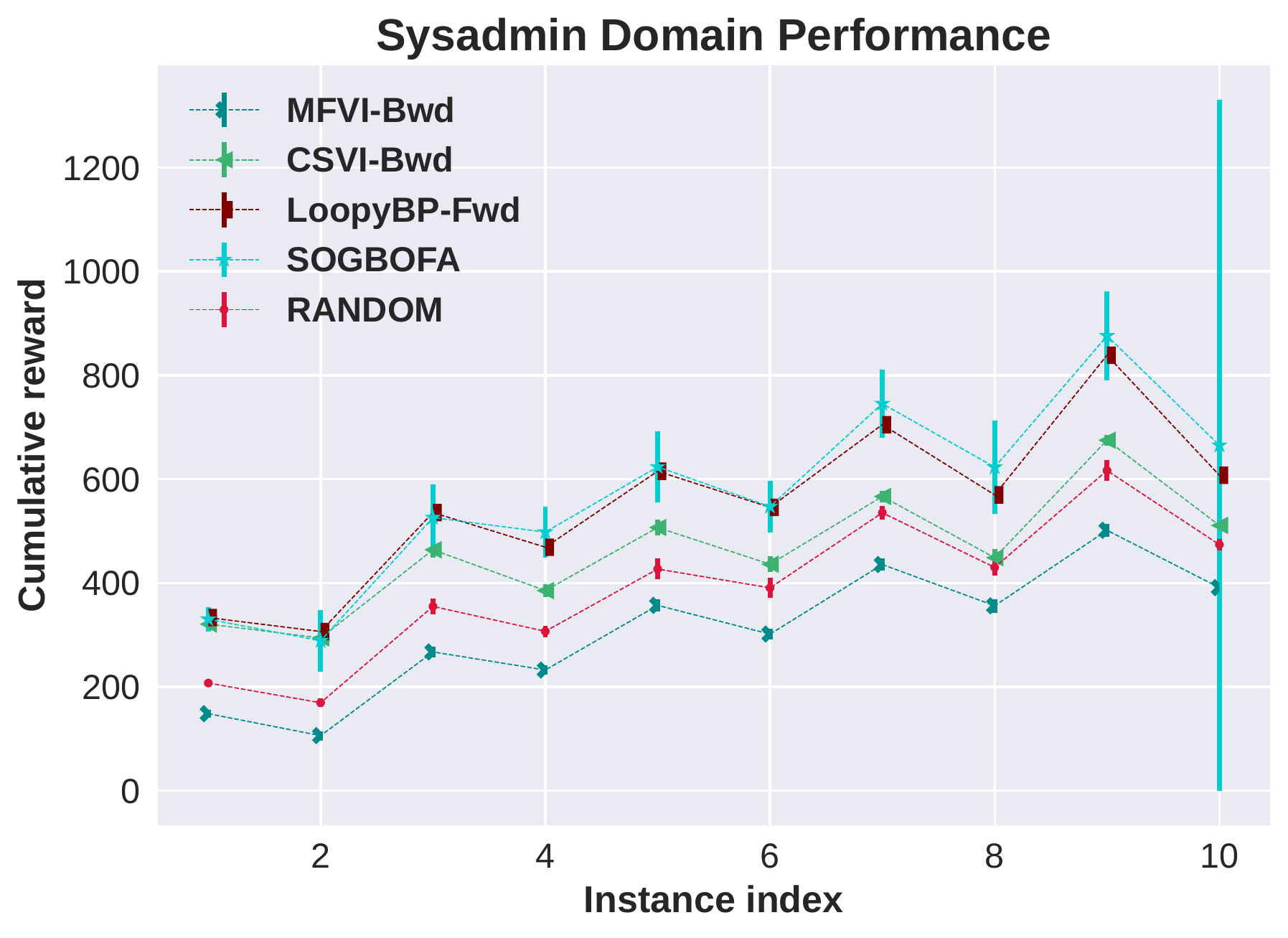}
	\end{subfigure}		
	\begin{subfigure}[t]{0.3\linewidth}
		\centering
		\includegraphics[width=\linewidth, trim=7 7 7 7, clip]{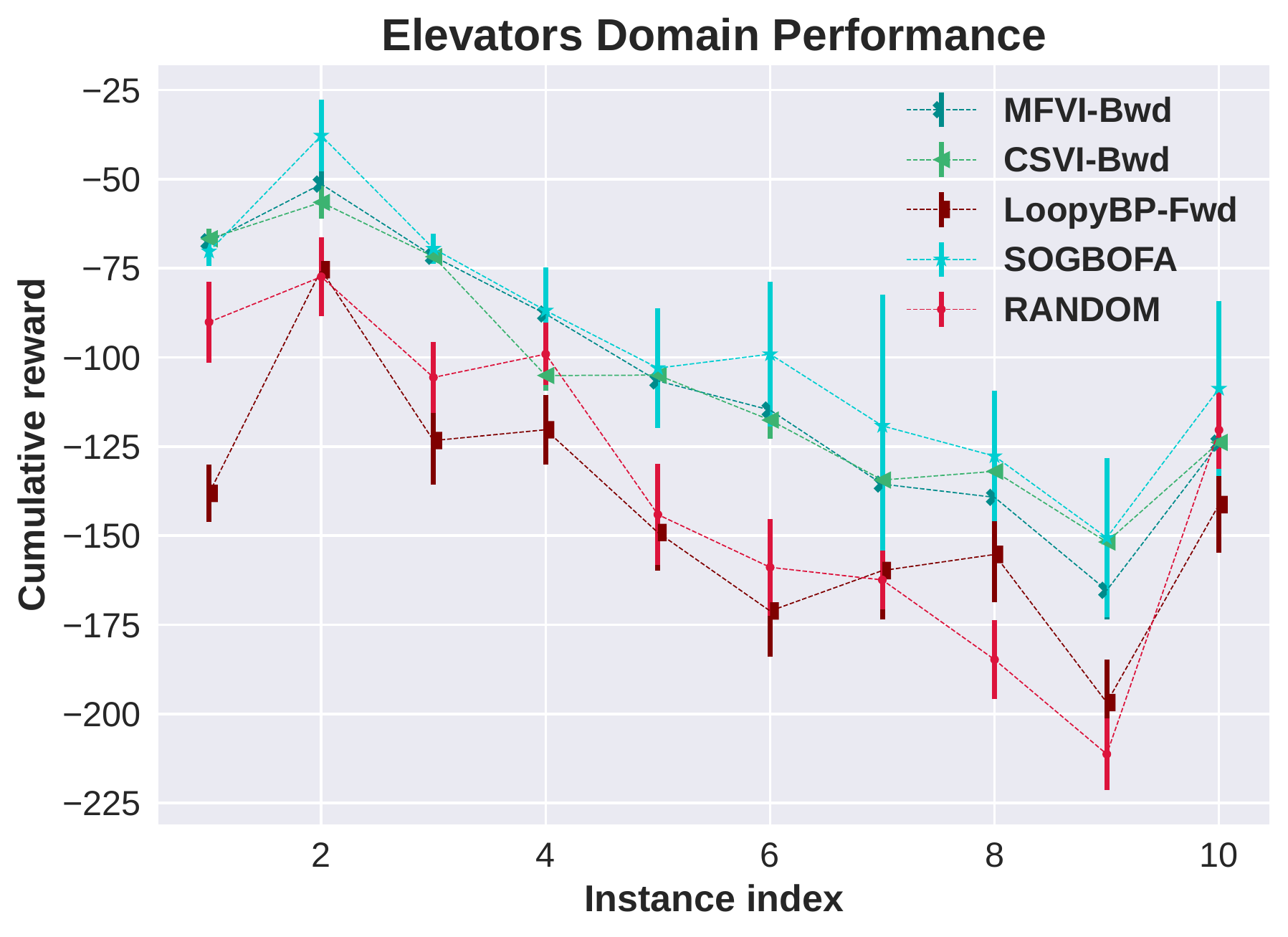}
	\end{subfigure}	
	\begin{subfigure}[t]{0.3\linewidth}
		\centering
		\includegraphics[width=\linewidth, trim=7 7 7 7, clip]{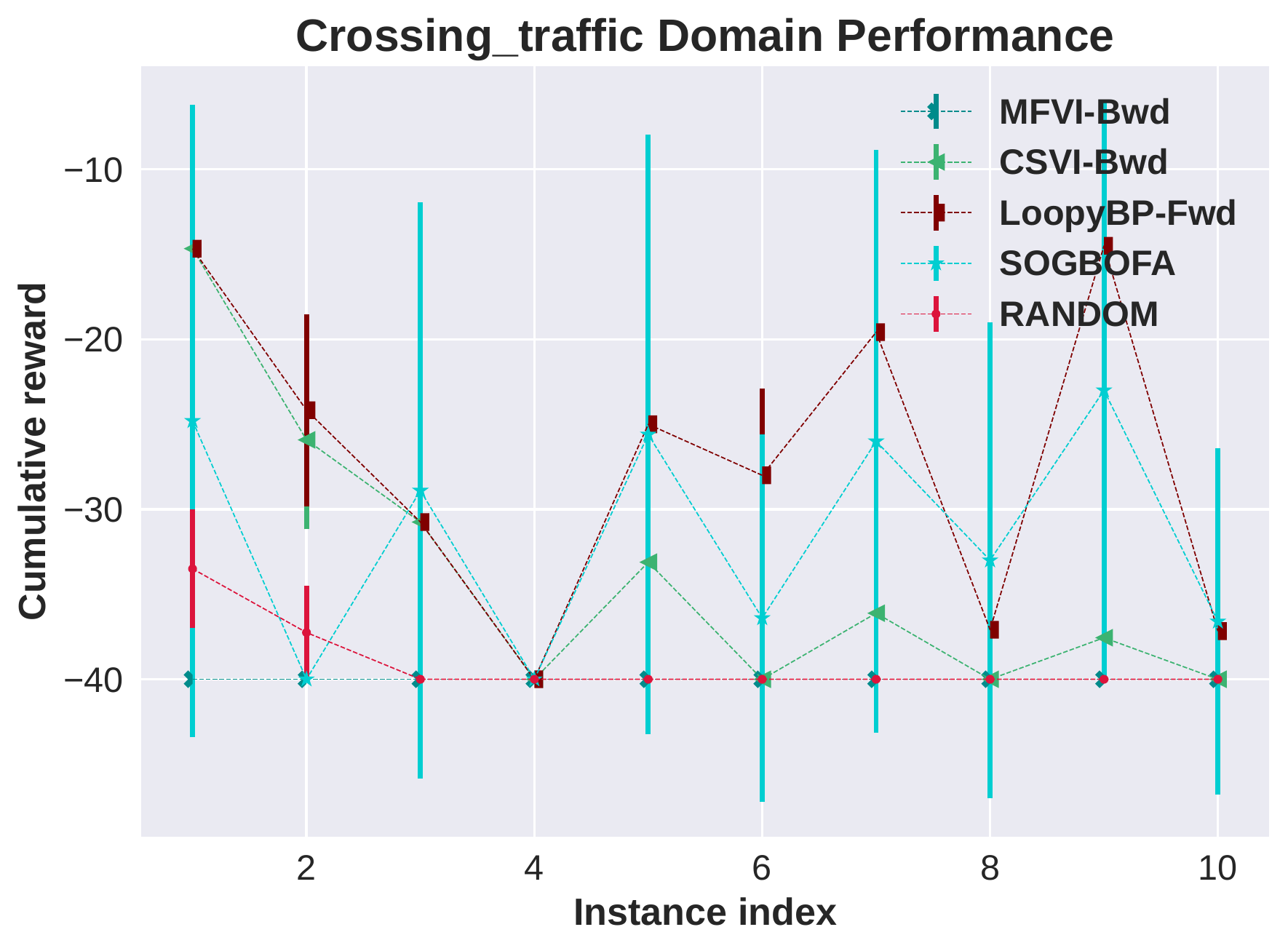}
	\end{subfigure}
	\begin{subfigure}[t]{0.3\linewidth}
		\centering
		\includegraphics[width=\linewidth, trim=7 7 7 7, clip]{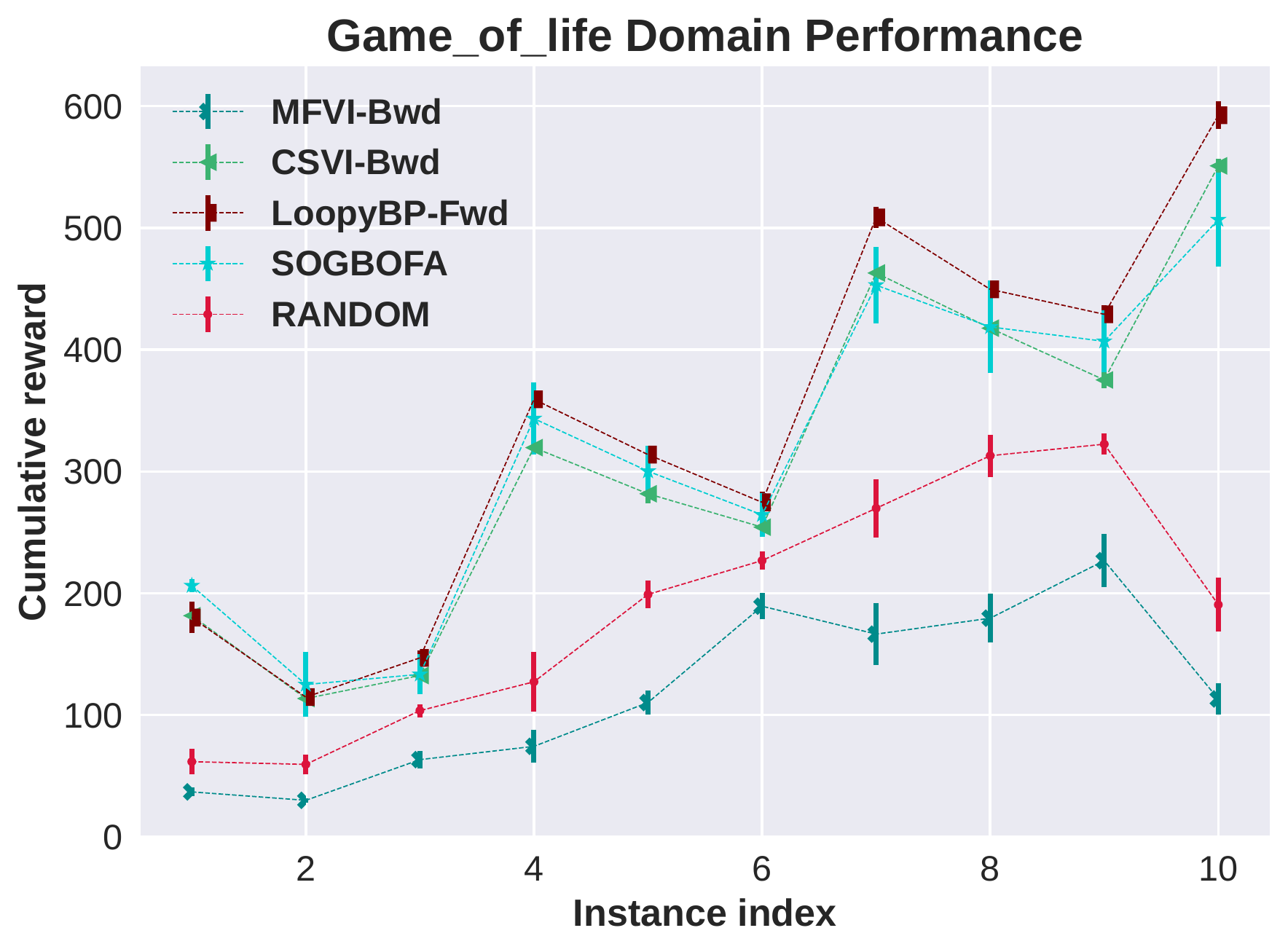}
	\end{subfigure}
	\begin{subfigure}[t]{0.3\linewidth}
		\centering
		\includegraphics[width=\linewidth, trim=7 7 7 7, clip]{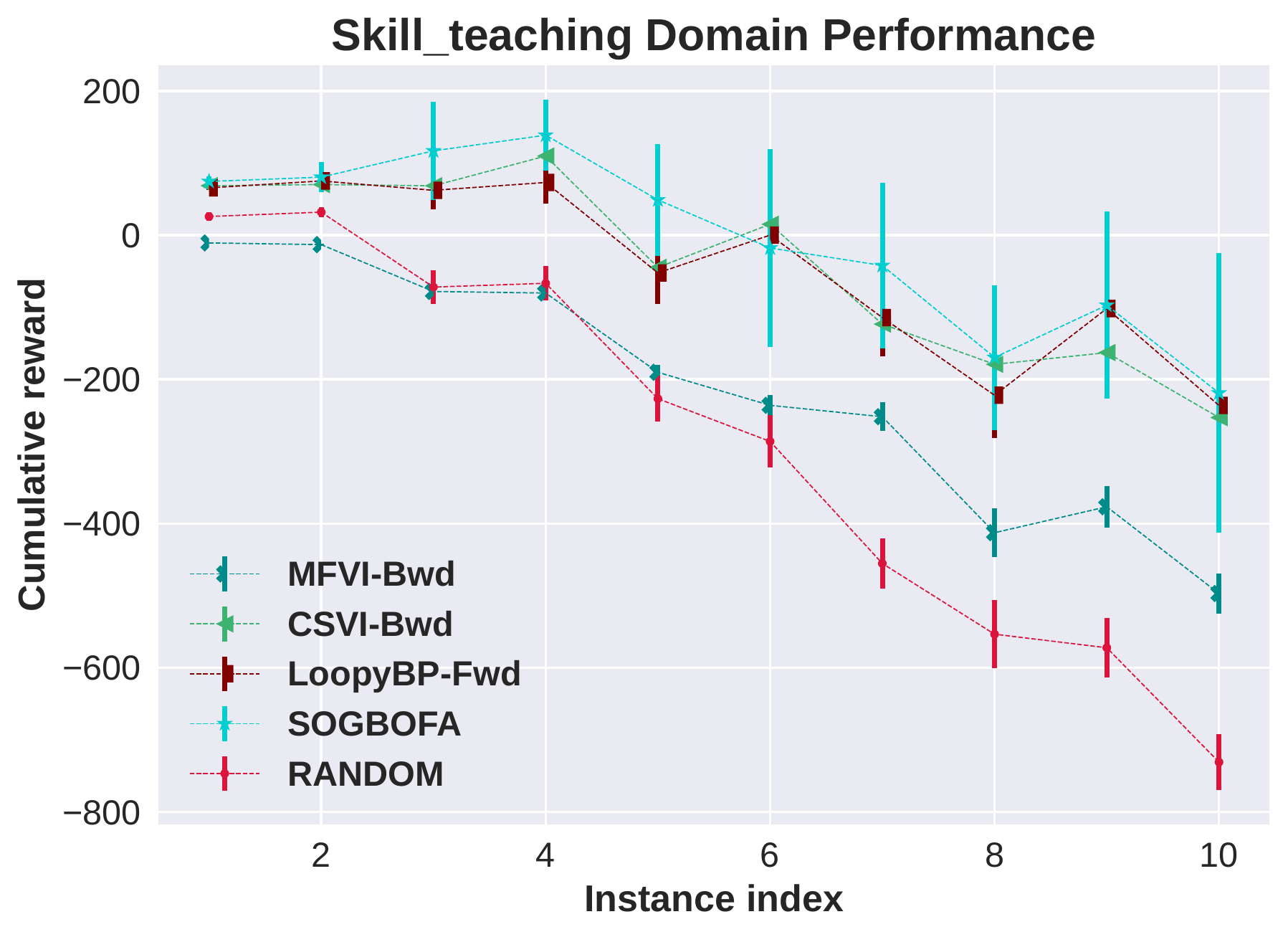}
	\end{subfigure}
	\begin{subfigure}[t]{0.3\linewidth}
		\centering
		\includegraphics[width=\linewidth, trim=7 7 7 7, clip]{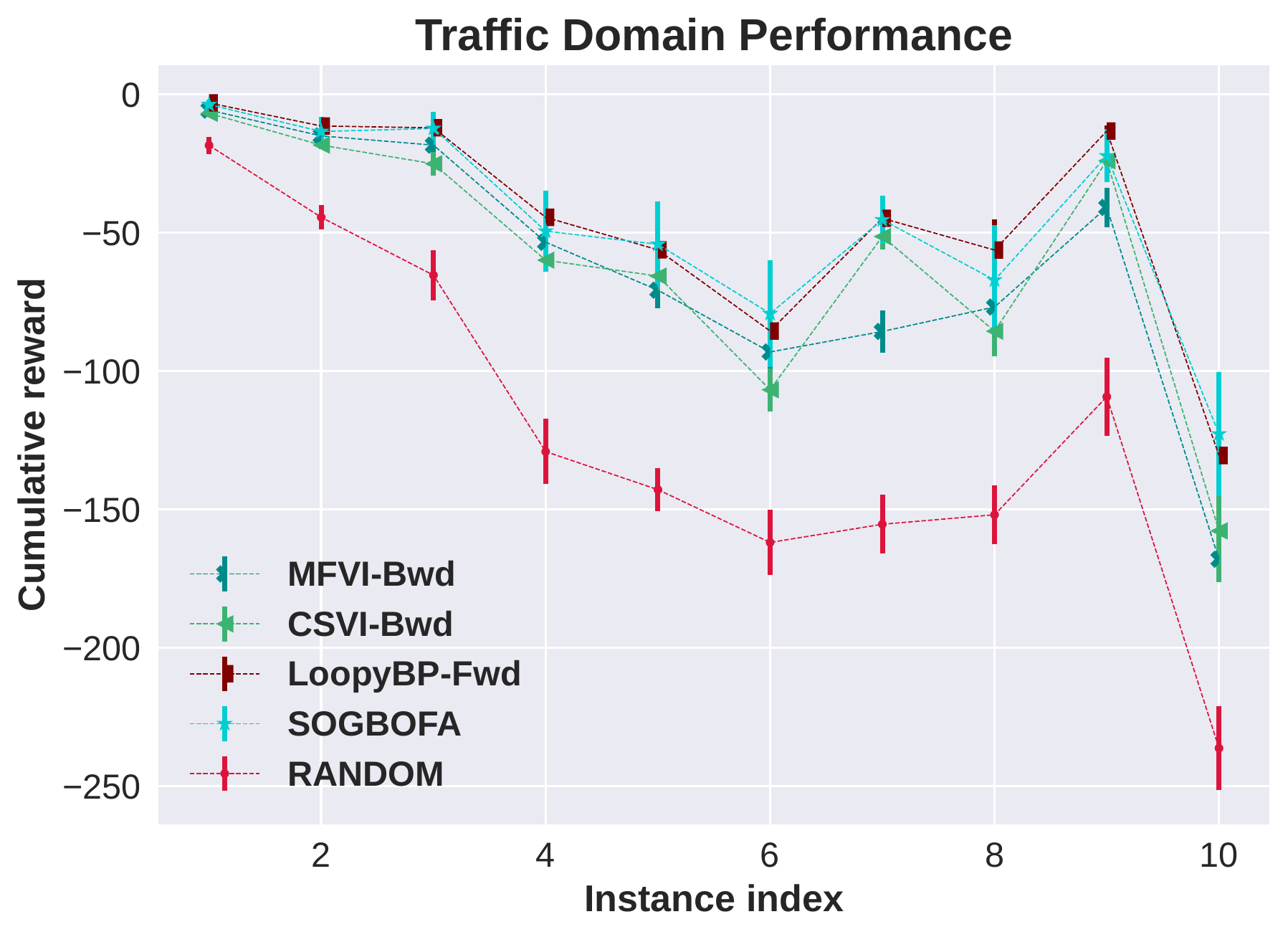}
	\end{subfigure}	
  \caption{Algorithm Performance Details}
\end{figure*}

\subsection{Different Optimization Strategies for Forward Loopy BP}
Recall that in the forward Loopy BP algorithm, we define $sc(\theta)$ to be the approximate marginal distribution of $p_{\theta}(c_T)$ computed by LBP. 
For the same construction, but using BP with a directed model, \cite{cui2018stochastic} showed that LBP does converge and that it does so in one iteration.
This holds because factors are conditional probability tables and we do not have downstream evidence.
The same holds in our case, for the corresponding message order.
However, this does not solve the optimization problem, i.e. selecting $\theta$ or $A$.

In this context we experimented with two methods. 
The first one is discussed in the main paper. Namely, the SOGBOFA algorithm \citep{cui2019stochastic} that fully optimizes $\theta$ by combining the one pass inference of the marginal problem, which is done symbolically, with a gradient search. 
The second one (labeled "LoopyBP-Fwd" in the plot) is
a computationally cheap compromise, introduced by \citet{cui2015factored}, which uses a uniform distribution for $a_1,\ldots,a_{T-1}$, 
and performs the optimization by enumerating values for $a_0$. That is, the second variant only optimizes the current action and uses a random rollout for subsequent actions. 
To make this as close as possible to SOGBOFA we used an implementation of BP with sequential updates where we can perform just one pass of forward messages (due to convergence).\footnote{
We have also experimented with the fast implementation using parallel updates as used by the backward algorithm which gives comparable results with 100 iterations of message propagation. 
} 

From the domain-by-domain experimental results we see that in all domains except Elevators the two algorithms have comparable and consistent performance. The need for optimizing the rollout policy for Elevators was discussed in \citet{cui2015factored} \citet{cui2019stochastic}. Briefly, a combination of positive and negative rewards in this domain means that random rollouts are not informative and, due to the large penalty, all actions look risky and the simpler planning algorithm chooses to do nothing. 
More importantly, for our experiments, the consistency in performance shows that the experimental advantage of SOGBOFA across all domains is not due to differences in implementation details but rather due to the inference strategy. 

\end{document}